\documentclass[11pt,leqno]{siamltex}
\usepackage[top=1in, bottom=1in, left=1in, right=1in]{geometry}
\usepackage{mathrsfs} 
\usepackage{amsfonts}
\usepackage{amsmath}
\usepackage{amssymb}
\usepackage{bm}
\usepackage{array}
\usepackage{subfigure}
\usepackage{graphicx}
\usepackage{algorithm} 
\usepackage{algpseudocode}
\usepackage{multirow}
\usepackage{bbm}
\usepackage{xcolor}
\usepackage{booktabs}
\usepackage{yfonts}
\usepackage{enumitem}
\usepackage{adjustbox}
\usepackage{mathtools}
\usepackage[left,mathlines]{lineno}

\newcommand{\R}{\ifmmode\mathbb{R}\else$\mathbb{R}$\fi}
\newcommand{\C}{\ifmmode\mathbb{C}\else$\mathbb{C}$\fi}
\newcommand{\N}{\ifmmode\mathbb{N}\else$\mathbb{N}$\fi}
\newcommand{\Q}{\ifmmode\mathbb{Q}\else$\mathbb{Q}$\fi}
\newcommand{\Z}{\ifmmode\mathbb{Z}\else$\mathbb{Z}$\fi}

\usepackage{hyperref}

\title{Identifying  Stochastic Dynamics via Finite expression methods  }

\author{
    Senwei Liang\\
    Lawrence Berkeley National Laboratory, Berkeley, CA 94720\\
    {\tt SenweiLiang@lbl.gov}
    \vspace{0.1in} \\
    Chunmei Wang\\
    Department of Mathematics, University of Florida,\\
    1400 Stadium Rd, Gainesville, FL 32611, USA\\
    {\tt chunmei.wang@ufl.edu}\vspace{0.1in} \\
    Xingjian Xu\\
    Department of Mathematics, University of Florida,\\
    1400 Stadium Rd, Gainesville, FL 32611, USA\\
    {\tt xingjianxu@ufl.edu}
}

\begin{document}
\maketitle
\begin{abstract}  Modeling stochastic differential equations (SDEs) is crucial for understanding complex dynamical systems in various scientific fields. Recent methods often employ neural network-based models, which typically represent SDEs through a combination of deterministic and stochastic terms. However, these models usually lack interpretability and have difficulty in generalizing beyond their training domain. This paper introduces the Finite Expression Method (FEX), a symbolic learning approach designed to derive interpretable mathematical representations of the deterministic component of SDEs. For the stochastic component, we integrate FEX with advanced generative modeling techniques to provide a comprehensive representation of SDEs. The numerical experiments on linear, nonlinear, and multidimensional SDEs demonstrate that FEX generalizes well beyond the training domain and delivers more accurate long-term predictions compared to neural network-based methods. The symbolic expressions identified by FEX not only improve prediction accuracy but also offer valuable scientific insights into the underlying dynamics of the systems.\\

 \textbf{Key words.} finite expression method, FEX, stochastic differential equations, symbolic learning, complex dynamical systems, interpretability
\end{abstract}

\section{Introduction}

A stochastic system is a dynamical system whose evolution is influenced by random processes or probabilistic rules, which results in inherently uncertain behavior. Modeling stochastic systems is important for uncovering the underlying mechanisms governing such randomness, which in turn enables understanding and prediction of their future states. This is of broad significance across numerous domains, with important applications in finance, biology, engineering, and physics~\cite{ bose2009stochastic,chong2020vortices,einstein1905molekularkinetischen,kariya2003options,rigas2015diffusive}.

Data-driven methods are widely applied to model stochastic systems using observed trajectory data~\cite{mou2023efficient}. For example, Gaussian processes \cite{archambeau2007gaussian,darcy2023one,opper2019variational} provide a nonparametric framework to capture uncertainty in functional relationships, while polynomial expansions \cite{li2021data,werner2024sample} offer a lightweight alternative with analytical tractability. These methods perform well in low-dimensional settings but become computationally prohibitive as dimensionality increases. In recent years, neural networks (NNs)~\cite{chen2023data,qin2019data} have emerged as a scalable alternative in scientific computing. Theoretical results show that NN approximations can alleviate the curse of dimensionality~\cite{montanelli2020error} and that optimization of over-parameterized NNs can achieve a low training error~\cite{du2019gradient}, making them a popular choice for modeling high-dimensional stochastic systems. In particular, generative models that incorporate NN architectures, such as autoencoders, generative adversarial networks \cite{opper2019variational,salimans2016improved} and diffusion models, have further advanced the ability to learn probabilistic representations and generate high-dimensional distributions. Among the latest advances, the training-free conditional diffusion model \cite{liu2024training} builds on this line of work by enabling efficient score-based learning without the need for NN training, which presents a promising alternative within the broader class of generative methods. Despite their empirical success, many NN-based methods function as black boxes, and they mainly prioritize predictive accuracy over understanding the underlying dynamics. This lack of interpretability can obscure physical insights and hinder their utility in scientific discovery. Moreover, their performance often deteriorates outside the training domain, raising concerns about generalizability. 

Symbolic regression provides a compelling alternative to NN-based methods in data-driven modeling~\cite{cranmer2020discovering,forrest1993genetic,jiang2023symbolic,kamienny2022end,li2022transformer,udrescu2020ai,wei2024closed}. Symbolic regression is a methodology that seeks interpretable mathematical expressions that best fit the data. In addition, symbolic regression can be implemented using search-based approaches such as genetic programming~\cite{cao2023genetic,oh2023genetic} and heuristic search~\cite{forrest1993genetic,jiang2023symbolic}, or via large-scale supervised pretraining~\cite{kamienny2022end,li2022transformer,liu2024kan}, where models are trained on synthetic datasets to efficiently generate candidate symbolic expressions. A representative method is Sparse Identification of Nonlinear Dynamical Systems (SINDy)~\cite{brunton2016discovering,fasel2021sindy,zhang2019convergence}, which formulates the task as a sparse linear regression problem. Although effective in many scenarios, SINDy assumes that the correct functional form lies within a predefined dictionary of candidate functions and restricts expressions to linear combinations, limiting its ability to model systems with more complex operator interactions. Alternatively, recent work has introduced finite expression method (FEX), a framework for discovering closed-form solutions of differential equations from a finite set of operator functions. In contrast to SINDy, FEX allows hierarchical compositions of operators to offer greater flexibility to represent a broader class of functional forms. It leverages reinforcement learning to address the combinatorial optimization problem of searching for mathematical expression solutions, enabling effective exploration of a broader solution space and producing compact, interpretable formulations with improved generalization. FEX has been applied to a wide range of problems, including high-dimensional PDEs \cite{liang2022finite}, dynamical systems \cite{du2024learning}, and several domain-specific applications \cite{jiang2023finite,song2024finite}.

In this paper, we introduce a two-stage framework that synergistically combines the NN-based method and symbolic regression. This framework decomposes a stochastic system into two components: a deterministic drift component, which captures the average trend, and a stochastic diffusion component, which represents the inherent random fluctuations. This two-stage framework leverages the distinct advantages of different modeling paradigms. \textcircled{\small{1}} The first stage employs symbolic regression to uncover the explicit mathematical form of the deterministic component. Such expression not only improves interpretability, but also reveals underlying structures such as conservation laws, symmetries. Specifically, we demonstrate the concept using FEX, which supports the automatic discovery of mathematical expressions with more complex compositional structures. \textcircled{\small{2}} The second stage focuses on modeling the stochastic component, which embodies the residual noise or diffusion remaining after accounting for the learned deterministic drift. To capture the complex and often non-Gaussian nature of this residual noise, we utilize advanced NN-based generative models to characterize the distribution of this residual noise. Compared to single-stage NN-based generative methods that predict the next state from the current state~\cite{chen2024learning,chen2024modeling,liu2024training,xu2024modeling,yang2024pseudoreversible}, the proposed two-stage framework enhances interpretability and generalization by potentially recovering the true mathematical expression of the drift component through symbolic regression.
 Numerical experiments on various systems, including linear, nonlinear, and multidimensional cases, demonstrate the advantages of the proposed framework. Additionally, this framework remains highly flexible to allow integration with other symbolic regression and generative methods.

The remainder of this paper is organized as follows. Section~\ref{ProbSet} introduces the problem setup and the proposed two-stage framework. Section~\ref{DesFEX} provides an overview of FEX, followed by a discussion on integrating a generative model, the training-free diffusion model~\cite{liu2024training}, in Section~\ref{DesSZ}. Section~\ref{Numerical} demonstrates the effectiveness of the proposed framework through multiple numerical experiments. Finally, we conclude the paper in Section~\ref{sec:con}.

\section {Problem Setup}\label{ProbSet}
Let $\left\{\Omega, \mathcal{F}, \mathbb{P}, \{ \mathcal{F}_t \}_{t \in [0,T] }\right\}$ be a complete filtered probability space, where $\Omega$ is the sample space, $\mathcal{F}$ is a $\sigma$-algebra on $\Omega$, $\mathbb{P}$ is a probability measure, and \(\{\mathcal{F}_t\}_{t \in [0,T] }\) is a filtration satisfying right-continuous and $\mathcal{F}_{0}$ contains all $\mathbb{P}-$null sets. 
The interval $[0,T]$ represents the finite time horizon over which the stochastic system evolves. 
This space supports a $m$-dimensional Brownian motion $W_{t}$, which governs the stochasticity of the system but is not directly observable. Let $\mathbf{x}_t \in \mathbb{R}^d$ denote the state of the system at time $t$. Consider the following $d-$dimensional stochastic differential equation (SDE), which describes the evolution of the system:
\begin{equation} \label{SDE}
d\mathbf{x}_t = \mu(\mathbf{x}_t)dt+\sigma(\mathbf{x}_t)dW_t,
\end{equation}
where $\mu: \mathbb{R}^{d} \to \mathbb{R}^{d}$ is a drift function, and $\sigma: \mathbb{R}^{d} \to \mathbb{R}^{d\times m}$ is a diffusion function, both satisfying  the Lipschitz continuity and linear growth conditions to ensure the existence and uniqueness of the SDE solution. 

In this work, we are given several trajectories of the system governed by the unknown SDE~\eqref{SDE}, where the exact forms of the drift $\mu$ and diffusion $\sigma$ functions are not known. Our goal is to infer the underlying stochastic dynamics from these observed trajectories and use the learned model to predict future states of the system based on its current state.

In the proposed  two-stage framework, we consider that unknown system can be represented by:
\begin{equation}\label{FM1}
\mathbf{x}_{t+\Delta_t} = \mathbf{x}_{t} + \Delta_t\underbrace{\mathbf{D}(\mathbf{x}_{t})}_{\text{deterministic}} + \underbrace{\mathbf{S}(z)}_{\text{stochastic}}, 
\end{equation}
where  \(\Delta_t>0\) is a step size and \(z \sim \mathcal{N}(0, \mathbf{I}_d) \) is a random variable drawn from a standard multivariate normal distribution.  Equation \eqref{FM1} decomposes the dynamics into two distinct components: a deterministic part \( \mathbf{D}(\mathbf{x}_t) \), which captures the average trend of change in the absence of randomness, and a stochastic part \( \mathbf{S}(z) \), which accounts for residual noise. This modeling framework employs a two-stage process to characterize these components sequentially.

In the first stage, we seek to identify each component of \( \mathbf{D}(\mathbf{x}_t) \). Specifically, for the \(i\)-th component, we consider a loss functional \(\mathcal{L}^{(i)}\), defined on a function space consisting of some function $u: \mathbb{R}^d \to \mathbb{R}$%~$\mathcal{G}$
, as:
\begin{equation} \label{population_functional}
   \mathcal{L}^{(i)}(u) := \int_{\Omega}\left\|\frac{\mathbf{x}_{t+\Delta_t}^{(i)}-\mathbf{x}_{t}^{(i)}}{\Delta_t}-u(\mathbf{x}_t)\right\|^2\rho(\mathbf{x}_t^{(i)})d\mathbf{x}_t^{(i)},
\end{equation}
where $\rho$ is the probability density function. The minimizer of $\mathcal{L}^{(i)}$ over the function space %$\mathcal{G}$ 
is denoted by $\hat{\mathbf{D}}^{(i)}$, of the \(i\)-th entry of \( \hat{\mathbf{D}}\), which serves as an approximation to \(\mathbf{D}\). In this work, we will obtain $\hat{D}$ using FEX which considers %~$\mathcal{G}$ to be 
a function space composed of mathematical expressions with a finite number of operators, as detailed in Section \ref{DesFEX}.

In the second stage, $\hat{\mathbf{D}}$ is substituted into Equation \eqref{FM1} to define the residual noise:
\begin{equation}\label{residual}
\mathbf{x}_{t+\Delta_t}-\mathbf{x}_{t}-\Delta_t \hat{\mathbf{D}}(\mathbf{x}_{t}).
\end{equation}
 The goal of modeling \(\mathbf{S}(z)\) is to recover the underlying distribution of the residual noise by mapping the standard multivariate normal variable \(z\) to this distribution. A generative model is well-suited for this purpose. While various generative models could be employed, we will use a training-free conditional diffusion model~\cite{liu2024training}, which has demonstrated state-of-the-art performance in learning stochastic systems. Instead of relying on its conditional formulation, we adapt this model to a non-conditional version. This version assumes that the residual noise is independent of the state $\mathbf{x}$. The detailed methodology is presented in Section~\ref{DesSZ}.

\section{First Stage to Learn \( \mathbf{D}(\mathbf{x}_t) \)}\label{DesFEX}
\begin{figure}[htbp]
    \centering
    \includegraphics[width=1\linewidth]{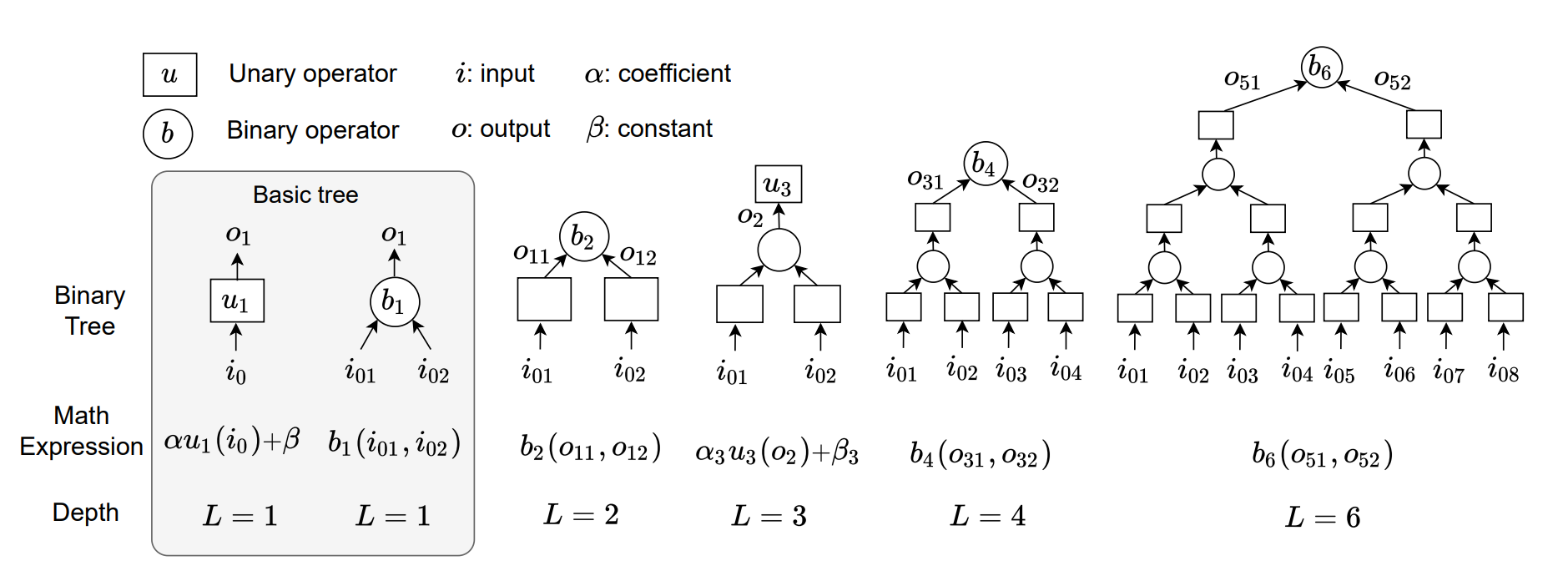}
    \caption{Computational rule of a binary tree. In a binary tree, each node is associated with either a unary or binary operator. We first describe the computation for a depth-1 tree (\(L = 1\)),
which contains only a single operator. For binary trees with more than depth one (\(L>1\)), the computation is performed recursively.}
    \label{binary_tree}
\end{figure}

We now present the details of the first stage in the two-stage framework to model the deterministic component in Equation \eqref{FM1}. In particular, FEX is employed to discover symbolic expressions represented as finite expressions within the function space \(\mathbb{S}\), which is spanned by a predefined set of operators. 
In FEX, a symbolic expression can be represented as a binary tree $\mathcal{T}$ with \(k\) nodes, each assigned either a unary or a binary operation, as illustrated in Figure~\ref{binary_tree}. The collections of unary operators and binary operators are denoted by $\mathbb{U}$ and $\mathbb{B}$, respectively, and are typically user-defined. Examples of unary and binary operators include:
\[\mathbb{U} = \left\{\sin,\exp,\log,(\cdot),(\cdot)^2,\frac{\partial}{\partial x_i},\cdots\right\} \text{ and } \mathbb{B} = \left\{+,-,\times,\div,\otimes,...\right\},\]
where each unary operator acts elementwise on the input and is typically implemented with learnable weights and bias to enhance its expressiveness.

Thus, for each output index \(i\)  of a state \(\bm{x}\in \mathbb{R}^d\), we collected an operator sequence \(\mathbf{e}\) from all nodes of $\mathcal{T}$. Combined with a set of trainable parameters \(\theta_1\), this forms a symbolic expression as \(u(\bm{x},\mathcal{T},\mathbf{e},\theta_1)\), where \(\bm{x}\) is an input and the output size is 1. In FEX, we set a function space
\[\mathbb{S}_{\text{FEX}} = \left\{u(\bm{x},\mathcal{T},\mathbf{e},\theta_1)\big| \mathbf{e}\in \mathbb{U}\cup \mathbb{B}, \ \text{for } i=1,\cdots,d\right\},\]
where \(\mathbb{S}_{\text{FEX}} \subset \mathbb{S}\) is used to seek a minimizer of the combinatorial optimization (CO) problem for a given binary tree \(\mathcal{T}\):
\begin{equation} \label{CO}
 \min \left\{{\mathcal{L}^{(i)}}(u(\cdot,\mathcal{T},\mathbf{e},\theta_1))| \mathbf{e}, \theta_1\right\}, \ \ \ i=1,\cdots,d.
\end{equation}

\subsection{Implementation of FEX} 
\begin{figure}[!ht]
    \centering
    \includegraphics[width=1\linewidth]{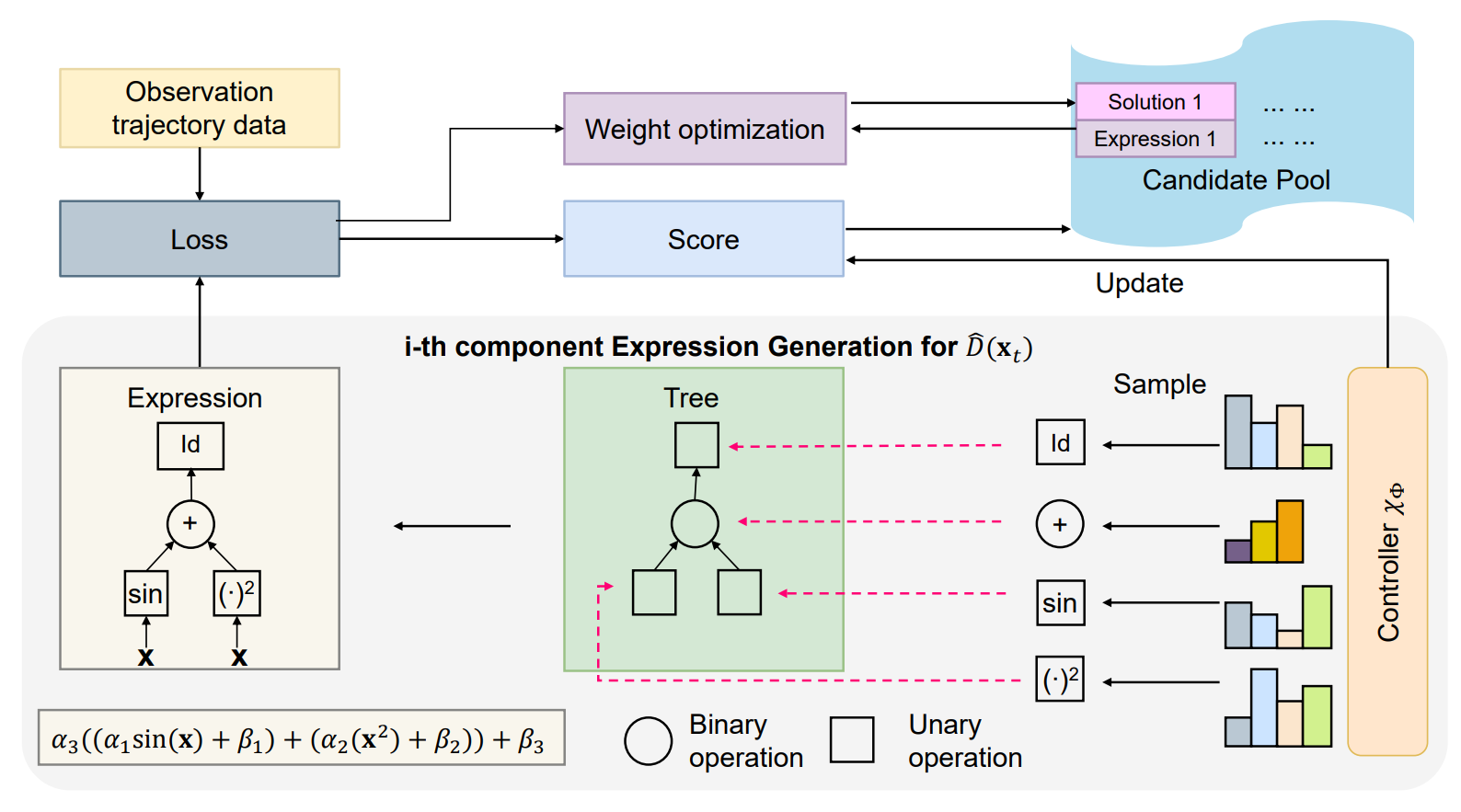}
    \caption{Representation of a search loop of FEX implementation.}
    \label{fig:FEX}
\end{figure}

The overall implementation of FEX, as shown in Figure \ref{fig:FEX}, is driven by a search loop based on reinforcement learning. This strategy is structured around four key components, which are detailed below. For simplicity, we restrict our presentation to the case \(d=1\), where the operator sequence is denoted by \(\mathbf{e}\) and the corresponding trainable parameters by \(\theta_1\).

\subsubsection{Score Computation}
Each operator sequence $\mathbf{e}$ is evaluated using a score function, denoted by $\text{Score}(\mathbf{e})$, defined as:
\[\text{Score}(\mathbf{e}) = \left(1+\sqrt{L(\mathbf{e})}\right)^{-1}, \quad \text{where} \quad L(\mathbf{e}) = \min\left\{{\mathcal{L}}(u(\cdot, \mathcal{T},\mathbf{e},\theta_1))|\theta_1\right\}. \]
Here, \(\text{Score}(\mathbf{e})\in [0,1]\). When \(L(\mathbf{e})\) approaches zero, the symbolic expression defined by \(\mathbf{e}\) closely approximates the true solution, and the score tends toward 1. 

To efficiently evaluate \(\text{Score}(\mathbf{e})\) globally, a two-step optimization approach is utilized here. An initial estimate is first obtained using a first-order method, such as stochastic gradient descent~\cite{gower2019sgd}, followed by refinement using the LBFGS algorithm \cite{fletcher2000practical,zhu1997algorithm}.
This yields a parameter set \(\theta_1^{\text{two-step}}\) and the score is approximated as
\[\text{Score}(\mathbf{e}) \approx (1+\mathcal{L}(u(\cdot,\mathcal{T},\mathbf{e},\theta_1^{\text{two-step}}))).\]

 \subsubsection{Operator Sequence Generation}
In FEX, a controller $\chi_{\Phi}$, parameterized by a set of learnable weights $\Phi$, generates a collection of probability mass functions $\{\mathbf{p}_{\Phi}^{j}\}_{j=1}^{k}$, which define the distributions over the \(k\) node values in \(\mathcal{T}\). An operator sequence \(\mathbf{e} = (e_1,e_2,\cdots,e_k)\) is then formed by sampling each \(\mathbf{e}_j\) from the corresponding distribution $\mathbf{p}_{\Phi}^{j}$. In addition, the $\epsilon-$ greedy strategy \cite{sutton2018reinforcement} is adopted to improve the exploration of a potentially high score $\mathbf{e}$ which can potentially improve the quality of the search process.

\subsubsection{Controller Update}
To encourage the generation of high-quality operator sequences, the controller parameters \(\Phi\) are updated to increase the likelihood of sampling promising candidates. This is achieved using a policy gradient approach to maximize the following objective function\cite{petersen2019deep}:
\begin{equation}
    \mathcal{J}(\Phi) = \mathbb{E}_{\mathbf{e}\sim \chi_{\Phi}}\left\{\text{Score}(\mathbf{e})|\text{Score}(\mathbf{e})\ge \text{Score}_{v,\Phi}\right\},
\end{equation}
where $\text{Score}_{v,\Phi}$ represents the $(1-v)\times 100\%$-quantile of the score distribution under $\chi_{\Phi}$. In other words, only sequences ranked within the top \(v\)-fraction are used to compute the gradient. In practice, the controller parameters are updated via thegradient descent with a learning rate $\eta$:
\[\Phi \leftarrow \Phi+ \eta \nabla_{\Phi}\mathcal{J}(\Phi).\]

\subsubsection{Candidate Optimization}
The FEX framework maintains a candidate pool $\mathcal{C}$, of size \(N\) to store high-scoring operator sequences $\mathbf{e}$ identified during the search. After completion of the search process, each sequence in $\mathcal{C}$ undergoes further refinement using an additional first-order optimization method with a small learning rate. This post-processing step helps reduce the risk of convergence to poor local minima by fine-tuning parameters around promising solutions.

\section{Second Stage to Learn \(\mathbf{S}(z)\)}\label{DesSZ}
We now present the second stage of the proposed framework to model the stochastic term in \eqref{FM1}. As an illustrative example, the model proposed by \cite{liu2024training} is designed for this purpose. It is implemented as a fully connected NN \(\hat{\mathbf{S}}_{\theta_2}(z)\) equipped with a parameter set $\theta_2$, and an activation function, such as tanh or ReLU. And this model is trained using specially constructed input–target pairs to identify a special decode transformation from the normal Gaussian distribution. 

\subsection{Construction of Data Pairs}\label{Section4.1}
A forward–reverse SDE technique is adopted over time \([0,1]\) here. 
The forward dynamics are defined as:
\begin{equation} \label{forwardSDE}
   dZ_{\tau} = b(\tau)Z_{\tau}d\tau+\sigma(\tau)dW_{\tau} \quad \text{with} \quad Z_0 = \mathbf{x}_{t+\Delta_t}-\mathbf{x}_{t}-\Delta_t \hat{\mathbf{D}}(\mathbf{x}_{t}),
\end{equation}
and the corresponding reverse SDEs are given as:
\begin{equation}\label{reverseSDE}
    dZ_{\tau} = \left[b(\tau)Z_{\tau}-\sigma^2(\tau)V(Z_{\tau},\tau)\right]d\tau+\sigma(\tau)dB_{\tau} \quad \text{with}\quad Z_1 = Z \sim \mathcal{N}(0,I_d),
\end{equation}
where $\tau \in [0,1)$, $V(Z_{\tau},\tau) =\nabla_Z \log p_{Z_{\tau}}(Z_{\tau})$, $B_\tau$ is the reverse-time Brownian motion, and \(I_d\) is a \(d\)-dimensional identity matrix. Here, \(b(\tau), \sigma(\tau)\) are scalar coefficient functions. Specifically, they are set as \[b(\tau) = \frac{d \log \alpha_{\tau}}{d\tau}, \ \ \ \ \sigma^2(\tau) = \frac{d\beta^2_{\tau}}{d\tau}-2\frac{d\log \alpha_{\tau}}{d\tau}\beta_{\tau}^2,\] with $\alpha_{\tau} = 1-\tau$, $\beta_{\tau}^2 = \tau$. This choice ensures that the forward SDE transports the distribution of residual noise \eqref{residual} toward normal Gaussian noise as $\tau \to 1$,  while the reverse SDEs reconstruct the samples by evolving the normal Gaussian noise back to the distribution of residual noise. 

By applying the reverse-time Fokker–Planck equation, along with the identity \(\nabla p_{Z_\tau}(Z_{\tau}) = p_{Z_{\tau}}(Z_{\tau})V(Z_{\tau},\tau)\), 
the reverse SDE be rewritten as the following ordinary differential equation (ODE): 
\begin{equation}\label{ODE}
   dZ_\tau = \left[b(\tau)Z_{\tau}-\frac{1}{2}\sigma^2(\tau)V(Z_{\tau},\tau)\right]d\tau, \ \text{with } Z_1 = Z \sim \mathcal{N}(0,I_d).
\end{equation}

Here, \cite{liu2024training} uses a Monte Carlo method to compute the value function \(V(Z_{\tau},\tau)\), instead of using neural networks in most of the common approaches. They use samples \(z \sim \mathcal{N}(0,I_d)\) as inputs and treat the numerical solution \(y\) of \eqref{ODE}
initialized at \(z\) as targets, thus forming supervised input-target pairs \({(z,y)}\). This setup enables the model to learn the reverse transformation that maps \(z\) to residuals.

\subsection{Optimization Problem}
In the second stage, \(\hat{\mathbf{S}}_{\theta_2}(z)\) is trained using the MSE loss between the predicted outputs and the target dataset in input-target pairs.  This leads to the optimization problem:
\begin{equation}\label{optim}
    \arg \min_{\theta_2}\mathbb{E}_{(z,y)}|| \hat{\mathbf{S}}_{\theta_2}(z)-y||_2^2.
\end{equation}
Once the optimal parameters $\theta^{*}$ are obtained by \eqref{optim} and applied to $\hat{\mathbf{S}}_{\theta_2}$, the resulting model
\(\hat{\mathbf{S}}(z) = \hat{\mathbf{S}}_{\theta^*}(z)\)
can be evaluated on any standard Gaussian sample to represent the learned stochastic component. Thus, this model is learned not only for the pointwise estimation but also for the distribution-level modeling, ensuring a more accurate representation of the stochastic process.

\section{Numerical Experiments}\label{Numerical} 

In this section, we present numerical experiments to demonstrate the effectiveness of the proposed two-stage framework on a variety of benchmark SDEs, as used in~\cite{liu2024training}. These examples aim to provide numerical evidence that:
\begin{enumerate}
    \item In the first stage, FEX is capable of accurately recovering the true drift function.
    \item  The framework can accurately estimate both drift and diffusion functions and also works well outside the training domain, showing strong generalizability. 
\end{enumerate}
In particular, we compare the proposed two-stage framework with the single-stage NN-based method introduced in \cite{liu2024training}, known as the training-free conditional diffusion method (TF-CDM). In our framework, we also adopt TF-CDM to model the noise component and thus refer to our method as FEX-DM. To evaluate the accuracy of the learned functions, we calculate the effective drift term and diffusion term given current state \(\bm{x}\) and predicted next state $\hat{\bm{x}}_{\Delta_t}$ as:
\begin{align}
\hat{\mu}(\bm{x}) &:= \mathbb{E}{\left[ \frac{\hat{\bm{x}}_{\Delta_t}-\bm{x}}{\Delta_t} \right]}, \ \ \ \ \ \
\hat{\sigma}(\bm{x}) :=\frac{\text{Std}(\hat{\bm{x}}_{\Delta_t})}{\sqrt{\Delta_t}}, 
\end{align}
where the expectation and standard deviation are estimated by multiple realization of $\hat{\bm{x}}_{\Delta_t}$. 

Then we implement the full FEX-DM workflow and summarize the experimental parameters below to support reproducibility.

\textbf{Data Generation.} For each SDE, we generate trajectory datasets \eqref{SDE} using the Euler–Maruyama method, with initial values uniformly sampled from predefined regions specific to each system. A total of $L$ trajectories are generated, each consisting of $N_{T}= 100$ time points obtained through uniform time discretization with a step size of $0.01$ over a total time horizon of $T=1.0$. Upon reorganizing these datasets into pairs \(\{(\mathbf{x}_t,\mathbf{x}_{t+\Delta_t})\}\), $L\times N_T$ samples are obtained. {As introduced in Section \ref{DesSZ}, we use the Euler method with a uniform time discretization of step size \(K = 10,000\) to get the numerical solution of \eqref{ODE} as the target dataset and adopt the same settings as in \cite{liu2024training} for the remaining parameters.}

\textbf{Setting of the First Stage.}
For each example, we choose the depth-3 tree with 3 unary operators and 1 binary operator (\(L=3\)) in Figure \ref{binary_tree} to generate closed-form mathematical expressions. The binary operator set is \( \mathbb{B} = \{+, -, \times\} \), and the unary operator set is \(\mathbb{U} = \{\sin, \cos, \exp, 0, \text{Id}, (\cdot)^2, (\cdot)^3, (\cdot)^4 \}.\) There are four main parts in the implementation of FEX as introduced in Section \ref{DesFEX}. We only briefly describe the key numerical setup here. \textcircled{\small{1}} \emph{Score Computation.} The score is optimized using the Adam with a learning rate $8.0\times 10^{-3}$ for $10,000$ iterations by solving \eqref{CO}. \textcircled{\small{2}} \emph{Controller Architecture.} A one-layer NN with ReLU is used here to represent the controller. Its output size is $|\mathbb{B}|+3|\mathbb{U}|$, where $|\cdot|$ denotes the cardinality of a set. \textcircled{\small{3}}  \emph{Controller Update.} The controller is updated using the policy gradient method with a batch size of 2, optimized by Adam with a learning rate $2.0\times 10^{-3}$. The number of training iterations depends on each example. \textcircled{\small{4}} \emph{Candidate Optimization.} In particular, we set $N=30$ to prevent incorrect expressions from replacing the correct ones within the training domain. For any $\mathbf{e} \in \mathbb{P}$, the parameters are optimized using Adam with an initial learning rate $5.0\times 10^{-3}$, decayed by a cosine decay schedule \cite{he2019bag} over $80,000$ iterations.

\textbf{Setting of the Second Stage.} For each example, a one-hidden-layer NN with 50 neurons and the tanh activation function is used for learning \(\mathbf{S}(z)\). This NN is trained for 2000 iterations using Adam with a learning rate $1.0\times 10^{-2}$ and a weight decay of $1.0\times10^{-6}$ to mitigate overfitting.

\subsection{Linear SDEs} We evaluate our framework on the one-dimensional Ornstein–Uhlenbeck (OU) process, a classic example with a well-characterized steady-state distribution that models noise-driven dynamics in finance \cite{vasicek1977equilibrium}, biology \cite{gardinerstochastic}, and statistical physics \cite{elgin1984fokker,uhlenbeck1930theory}. 
The evolution of the one-dimensional OU process is governed by:
\begin{equation} \label{OUprocess}
    d\mathbf{x}_t = \theta(\mu-\mathbf{x}_t)dt+\sigma dW_t,
\end{equation}
where $\theta = 1.0, \mu = 1.2$ and $\sigma = 0.3$. We simulate $L = 15,000$ trajectories with initial values uniformly sampled from $(0,2.5)$ to train the two-stage framework. After only 90 training iterations in the first stage, the controller NN identifies an operator sequence identical to that of the true drift function. Combined with candidate optimization, this generates the best score expression for $\hat{D}(\mathbf{x}_t)$, as shown in Table \ref{OUTab}.  It indicates that the deterministic component has been successfully captured by FEX.

\begin{figure}[!htbp]
    \centering
    \includegraphics[width=0.9\linewidth]{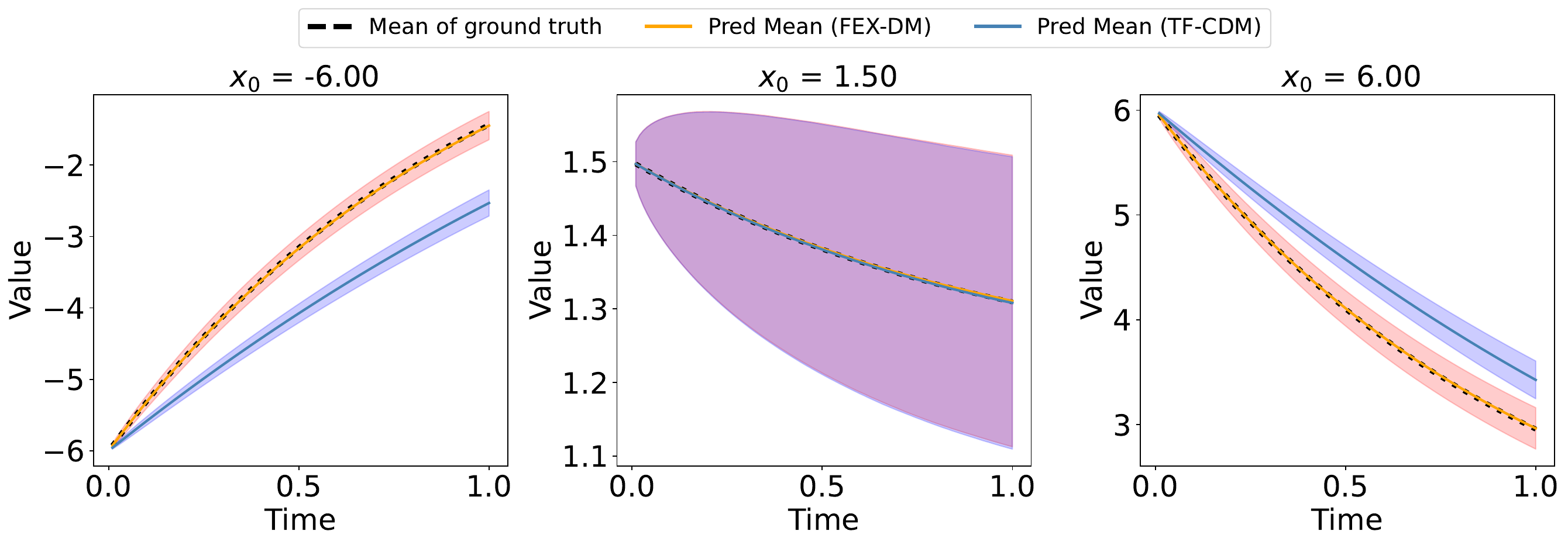}
    \caption{
    Comparison of predicted mean trajectories and corresponding mean $\pm$ standard deviation bands of solutions for 1-dimensional OU process, evaluated at initial conditions $\mathbf{x}_0=-6, 1.5, 6$, obtained using the FEX-DM, TF-CDM, and the ground truth.} \label{Fig1OU}
\end{figure}

\begin{figure}
    \centering
    \includegraphics[width=0.9\linewidth]{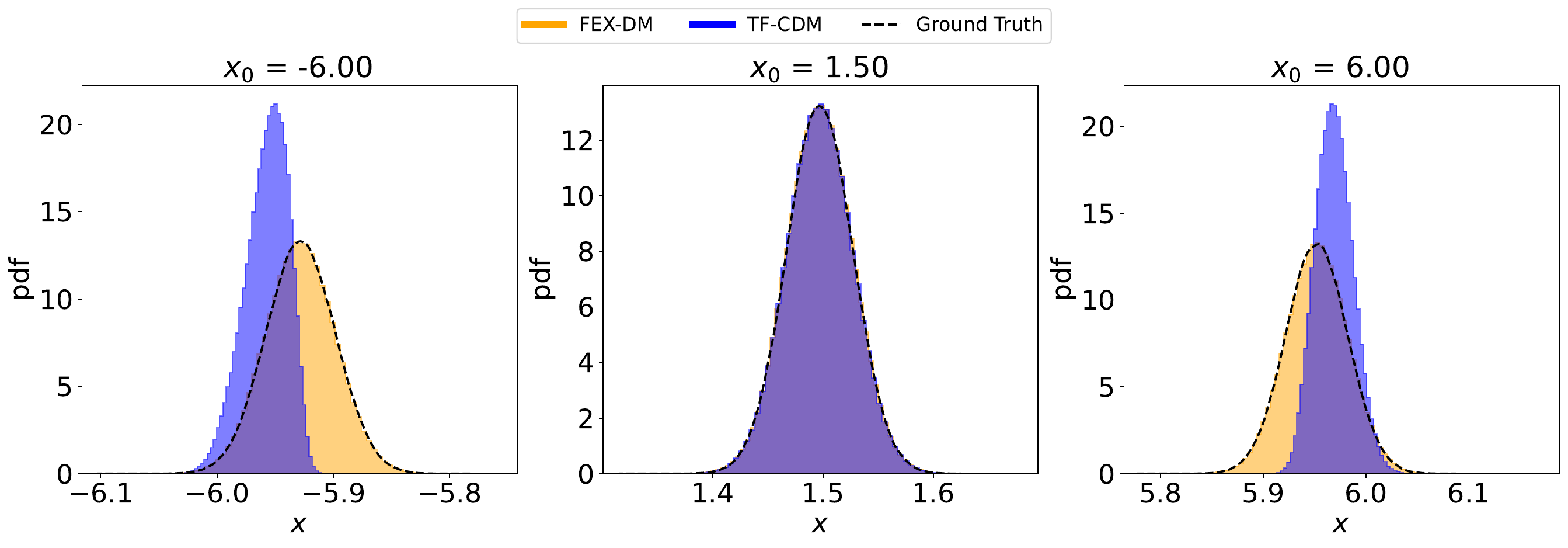}
    \caption{Comparison of conditional PDF $p_{\mathbf{x}_{t+\Delta t|\mathbf{x}_t}}(\mathbf{x}_{t+\Delta t}|\mathbf{x}_t = c)$ determined by FEX-DM, TF-CDM and the ground truth for 1-dimensional OU process, with $c=-6, 1.5, 6$ at $t= 0$,  respectively.}\label{Fig2OU}
\end{figure}

\begin{figure}[!htbp]
    \centering
    \includegraphics[width=0.66\linewidth]{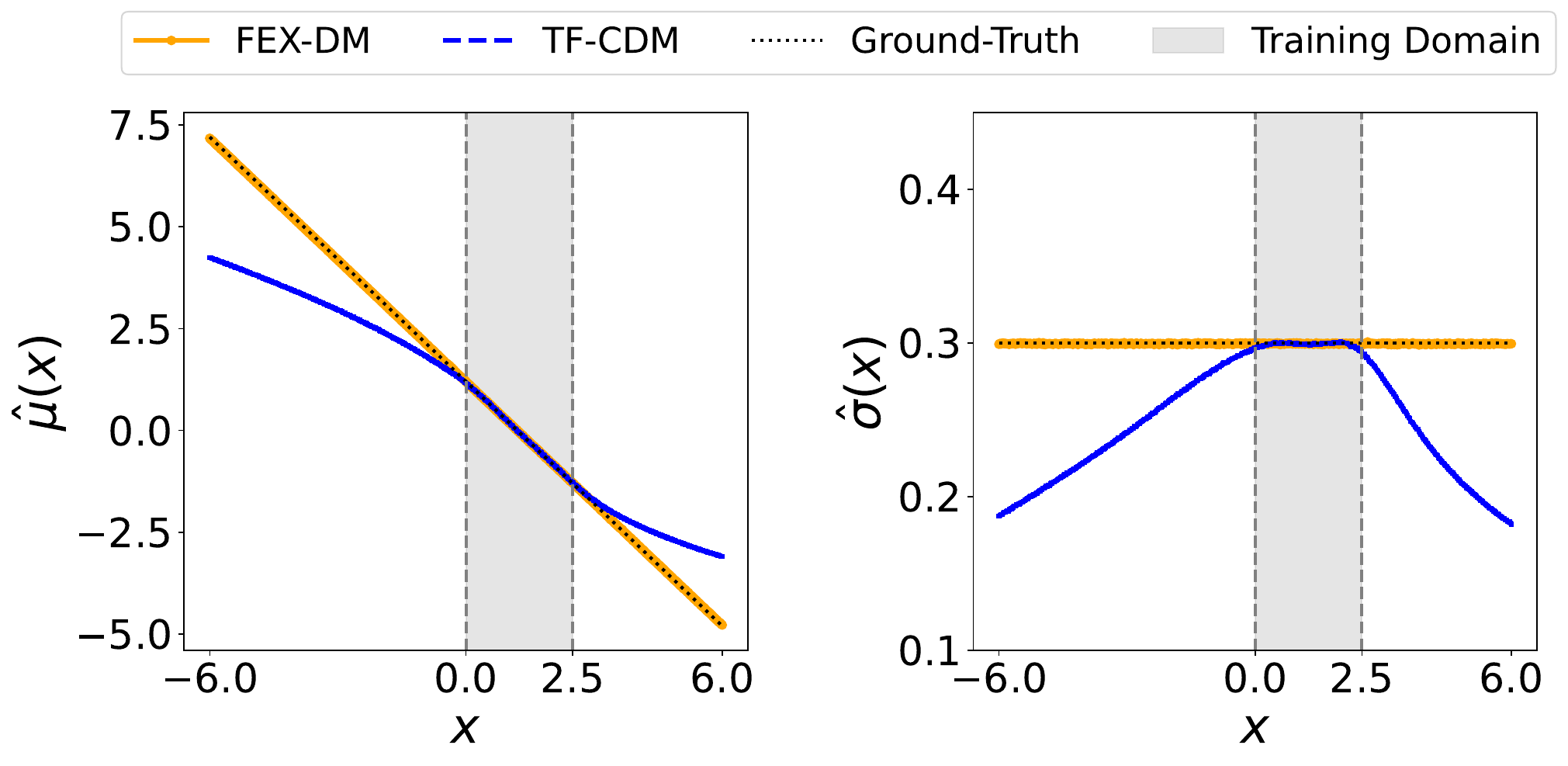}
    \caption{Comparison of drift and diffusion functions obtained by the FEX-DM, TF-CDM and the ground truth over domain $[-6,6]$ for 1-dimensional OU process. Left: drift term $\hat{\mu}(x)$; Right: diffusion term $\hat{\sigma}(x)$; The shaded region indicates the training domain $[0,2.5]$.}\label{Fig3OU}
\end{figure}

\begin{table}[!ht]
\centering
\caption{Comparison of true drift term and the FEX-learned deterministic component.} \label{OUTab}
\begin{tabular}{| >{\centering\arraybackslash}m{6cm} |>{\centering\arraybackslash}m{6cm} |}
\hline
\textbf{True drift expression} &
\textbf{Best expression for $\hat{D}(\mathbf{x}_t)$} \\ \hline
 $1.2 - \mathbf{x}_t$ &
$1.1989-0.9953\mathbf{x}_t$
 \\ \hline
\end{tabular}
\end{table}
After the second-stage training, we randomly select 100,000 trajectories with initial conditions $\mathbf{x}_0 = -6,1.5,6$, covering both in-domain and out-of-domain values to evaluate the framework's ability to predict SDE dynamics over \(T=1\).
Figure \ref{Fig1OU} compares the predicted mean trajectories and uncertainty bands, while Figure \ref{Fig2OU} shows the corresponding one-step conditional distributions. In both views, FEX-DM consistently aligns better with the ground truth than TF-CDM, accurately capturing mean, variance, and full conditional structure, even for out-of-domain inputs where TF-CDM deviates significantly in uncertainty and distributional shape. In addition, we test the effective drift \(\hat{\mu}\) and \(\hat{\sigma}\) estimated by our framework in the extended
domain \([-6,6]\), as shown in Figure \ref{Fig3OU}. This further demonstrates the strong generalizability of FEX-DM.

\subsection{Nonlinear SDEs} In this section, we consider three nonlinear SDE cases. The first models the periodic dynamics of biological rhythms, climate patterns, and fluid flows ~\cite{karatzas1991brownian, oksendal2013stochastic}. The second describes a double-well potential, challenging to model in chemical reaction kinetics, statistical mechanics, and phase transitions~\cite{hanggi1990reaction,kramers1940brownian}. The third is the Oscillatory Langevin (OL) process, presenting additional modeling difficulties in molecular dynamics, soft matter physics, and non-equilibrium thermodynamics~\cite{coffey2012langevin,gardiner2009stochastic}. 

\subsubsection{1-dimensional SDE with Trigonometric Drift} We evaluate our framework on the SDE with trigonometric drift function, given by
\begin{equation}\label{TriSDE}
    d\mathbf{x}_t = \sin(2m\pi\mathbf{x}_t)dt+\sigma dW_t,
\end{equation}
where $m=1$ and $\sigma=0.8$. A total of $10,000$ trajectories are simulated, with initial values uniformly sampled from $(0,1)$ for training. After only 120 iterations in the first stage, the controller NN identifies an operator sequence that aligns with the true drift structure up to trigonometric equivalence, capturing the correct functional form with symmetry and a phase-shifted cosine representation. Combined with candidate optimization, this generates the best score expression for $\hat{D}(\mathbf{x}_t)$, as shown in Table \ref{TriTab}.  It indicates that the deterministic component has been successfully captured by FEX.

\begin{table}[!ht]
\centering
\caption{Comparison of true drift term and the FEX-learned deterministic component.} \label{TriTab}
\begin{tabular}{| >{\centering\arraybackslash}m{6cm} |>{\centering\arraybackslash}m{6cm} |}
\hline
\textbf{True drift expression} &
\textbf{Best expression for $\hat{D}(\mathbf{x}_t)$} \\ \hline
 $\sin(2\pi\mathbf{x}_t)$ &
$-1.1989\cos(6.2476\mathbf{x}_t-4.6837)- 0.0104$
 \\ \hline
\end{tabular}

\end{table}

\begin{figure}[!htbp]
    \centering
    \includegraphics[width=0.9\linewidth]{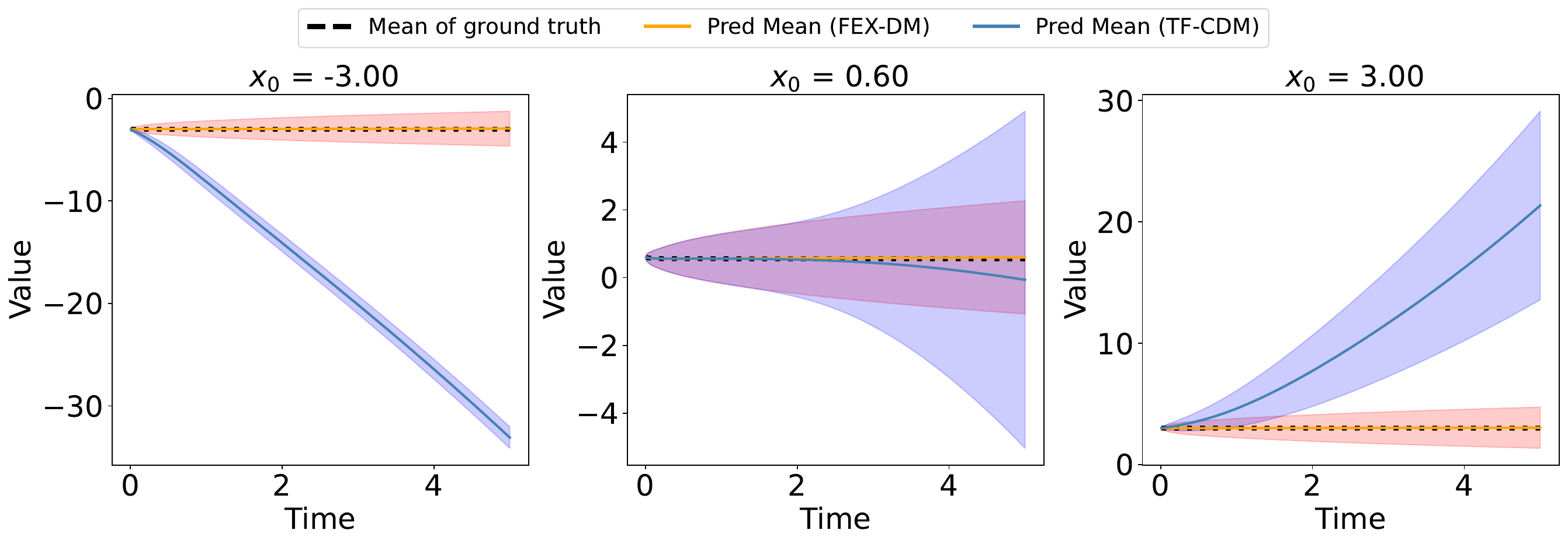}
    \caption{ Comparison of predicted mean trajectories and corresponding mean $\pm$ standard deviation bands of solutions for SDE with trigonometric drift, evaluated at initial conditions $\mathbf{x}_0=-3, 0.6, 3$, obtained using the FEX-DM, TF-CDM, and the ground truth.} \label{Fig1Tri}
\end{figure}

\begin{figure}[!htbp]
    \centering
    \includegraphics[width=0.9\linewidth]{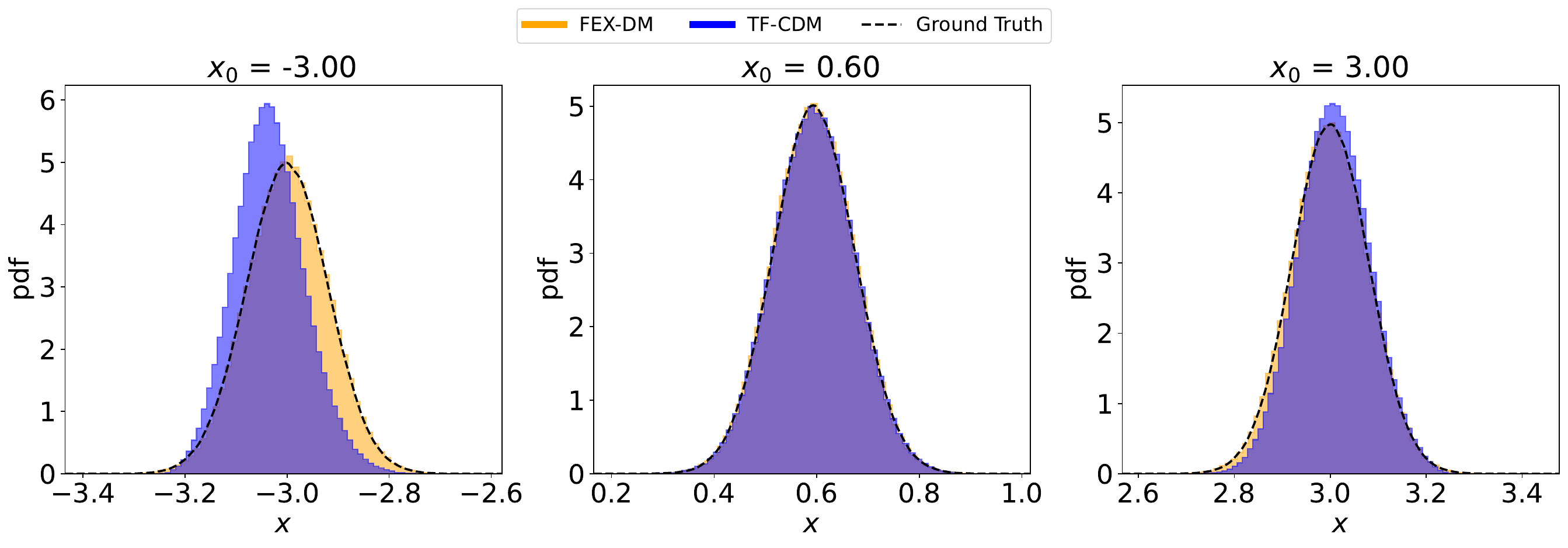}
    \caption{Comparison of conditional PDF $p_{\mathbf{x}_{t+\Delta t|\mathbf{x}_t}}(\mathbf{x}_{t+\Delta t}|\mathbf{x}_t = c)$ determined by FEX-DM, TF-CDM and the ground truth for SDE with trigonometric drift, with $c=-3, 0.6, 3$ at $t= 0$,  respectively.}\label{Fig2Tri}
\end{figure}

After the second-stage training, we randomly select 500,000 trajectories with initial conditions $\mathbf{x}_0 = -3,0.6,3$, covering both in-domain and out-of-domain values to evaluate the framework's ability to predict SDE dynamics over \(T=5\).
Figure \ref{Fig1Tri} compares the predicted mean trajectories and uncertainty bands, while Figure \ref{Fig2Tri} shows the corresponding one-step conditional distributions. In both views, FEX-DM consistently aligns better with the ground truth than TF-CDM, accurately capturing mean, variance, and full conditional structure, even for out-of-domain inputs where TF-CDM deviates significantly in uncertainty and distributional shape. In addition, we test the effective drift \(\hat{\mu}\) and \(\hat{\sigma}\) estimated by our framework in the extended
domain \([-5,5]\), as shown in Figure \ref{Fig3OU}. This further demonstrates the strong generalizability of FEX-DM.

\begin{figure}[!htbp]
    \centering
    \includegraphics[width=0.65\linewidth]{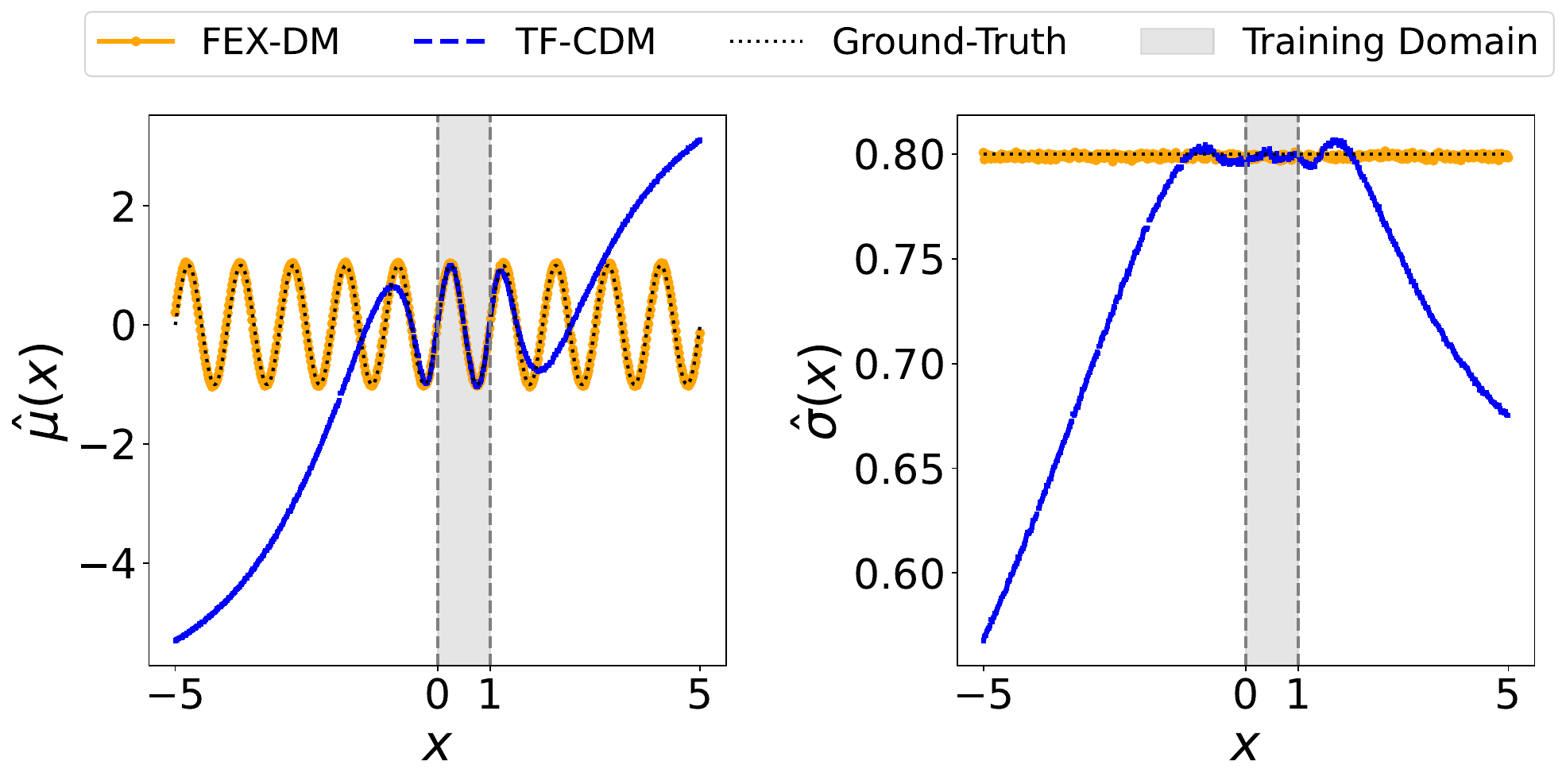}
    \caption{
    Comparison of drift and diffusion functions obtained by the FEX-DM, TF-CDM and the ground truth over domain $[-5,5]$ for SDE with trigonometric drift. Left: drift term $\hat{\mu}(x)$; Right: diffusion term $\hat{\sigma}(x)$; The shaded region indicates the training domain $[0,1]$.
    }\label{Fig3Tri}
\end{figure}

\subsubsection{1-dimensional SDE with Double-Well Potential} We test our framework on an SDE governed by a double-well potential:
\[dx_t = (x_t-x_t^3)dt+\sigma dW_t,\]
where $\sigma = 0.5$. This system has two stable states at $x=-1$ and $x=1$, and the trajectory can randomly switch between them over time. For data generation, $10,000$ trajectories are generated with the initial conditions uniformly sampled from $(-2,2)$.  After 500 training iterations in the first stage, the controller NN identifies an operator sequence identical to that of the true drift function. Combined with candidate optimization, this generates the best score expression for $\hat{D}(\mathbf{x}_t)$, as shown in Table \ref{DwellTab}.  It indicates that the deterministic component has been successfully captured by FEX. 

\begin{table}[!ht]
\centering
\caption{Comparison of true drift term and the FEX-learned deterministic component.} \label{DwellTab}
\begin{tabular}{| >{\centering\arraybackslash}m{6cm} |>{\centering\arraybackslash}m{6cm} |}
\hline
\textbf{True drift expression} &
\textbf{Best expression for $\hat{D}(\mathbf{x}_t)$} \\ \hline
 $\mathbf{x}_t-\mathbf{x}_{t}^3$ &
$-0.9922\mathbf{x}_{t}^3+0.9709\mathbf{x}_{t}+0.0019$
 \\ \hline
\end{tabular}

\end{table}

\begin{figure}[!htbp]
    \centering
    \includegraphics[width=0.9\linewidth]{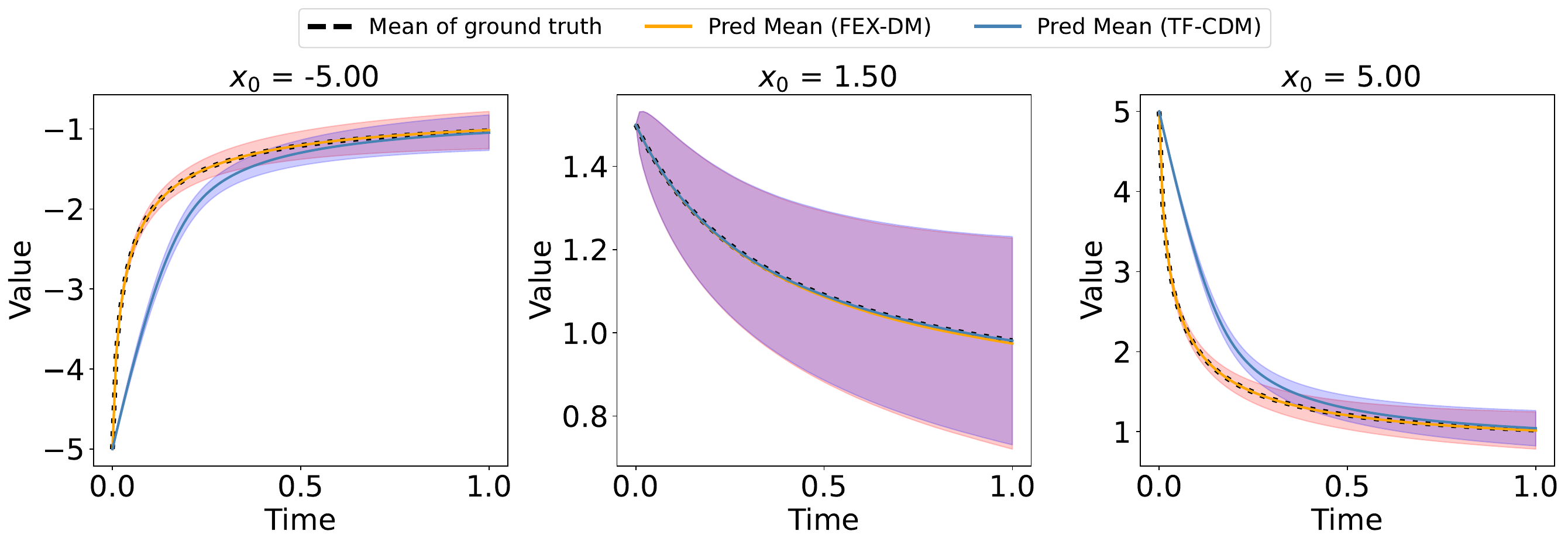}
    \caption{
    Comparison of predicted mean trajectories and corresponding mean $\pm$ standard deviation bands of solutions for SDE with double well potential, evaluated at initial conditions $\mathbf{x}_0=-5, 1.5, 5$, obtained using the FEX-DM, TF-CDM, and the ground truth.} \label{Fig1Dwell}
\end{figure}

\begin{figure}[!htbp]
    \centering
    \includegraphics[width=1\linewidth]{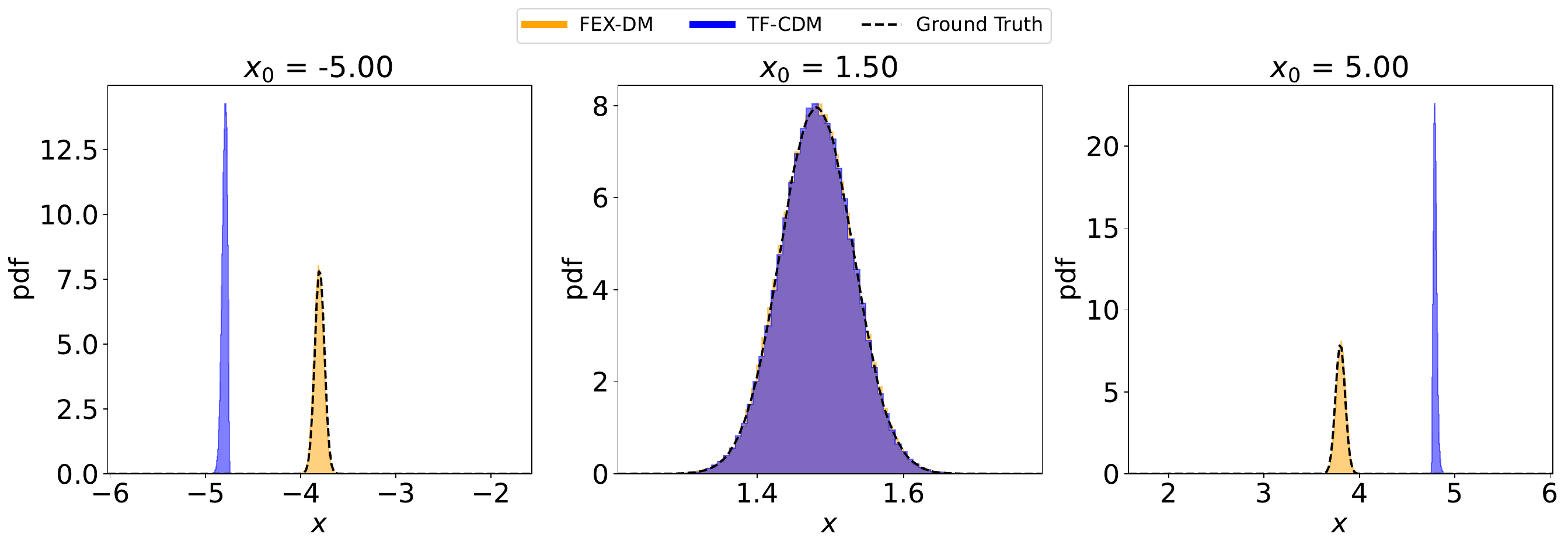}
    \caption{Comparison of conditional PDF $p_{\mathbf{x}_{t+\Delta t|\mathbf{x}_t}}(\mathbf{x}_{t+\Delta t}|\mathbf{x}_t = c)$ determined by FEX-DM, TF-CDM and the ground truth for SDE with double well potential, with $c=-5,1.5,5$ at $t=0$, respectively.
}\label{Fig2DwellExtended}
\end{figure}

\begin{figure}
    \centering
    \includegraphics[width=1\linewidth]{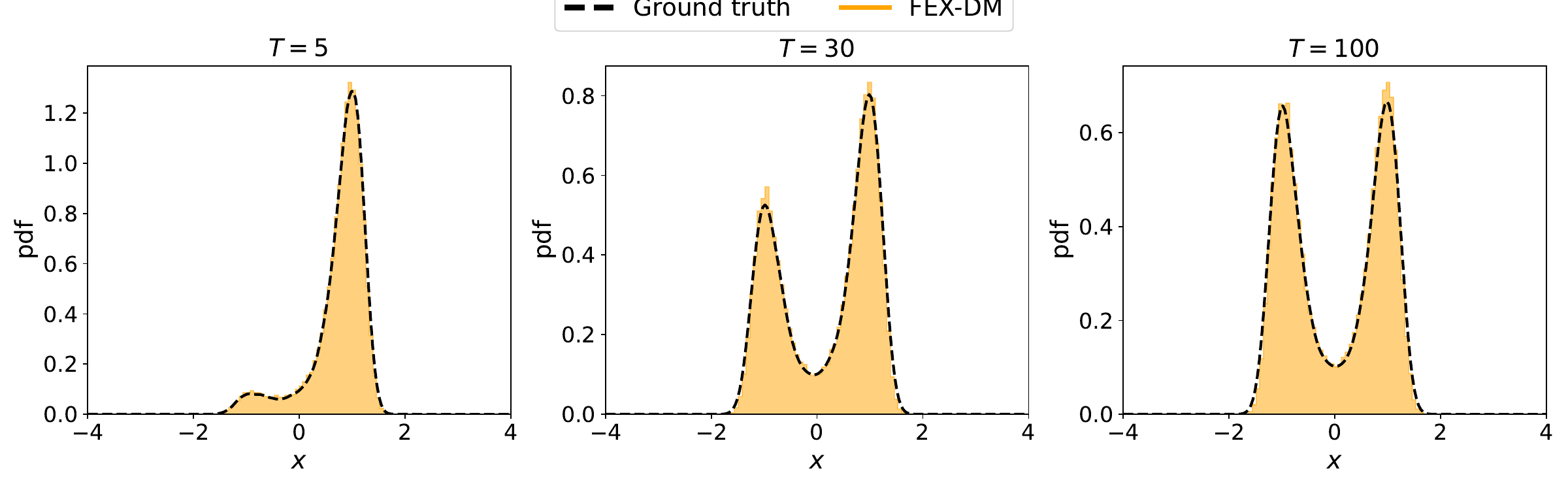}
    \caption{Evolution of conditional PDF $p_{\mathbf{x}_{t+\Delta t|\mathbf{x}_t}}(\mathbf{x}_{t+\Delta t}|\mathbf{x}_t = 1.5)$ determined by FEX-DM and the ground truth for SDE with double well potential, at $t=5, 30, 100$, respectively.}
    \label{fig:DwellEvul}
\end{figure}

\begin{figure}[!htbp]
    \centering
    \includegraphics[width=0.65\linewidth]{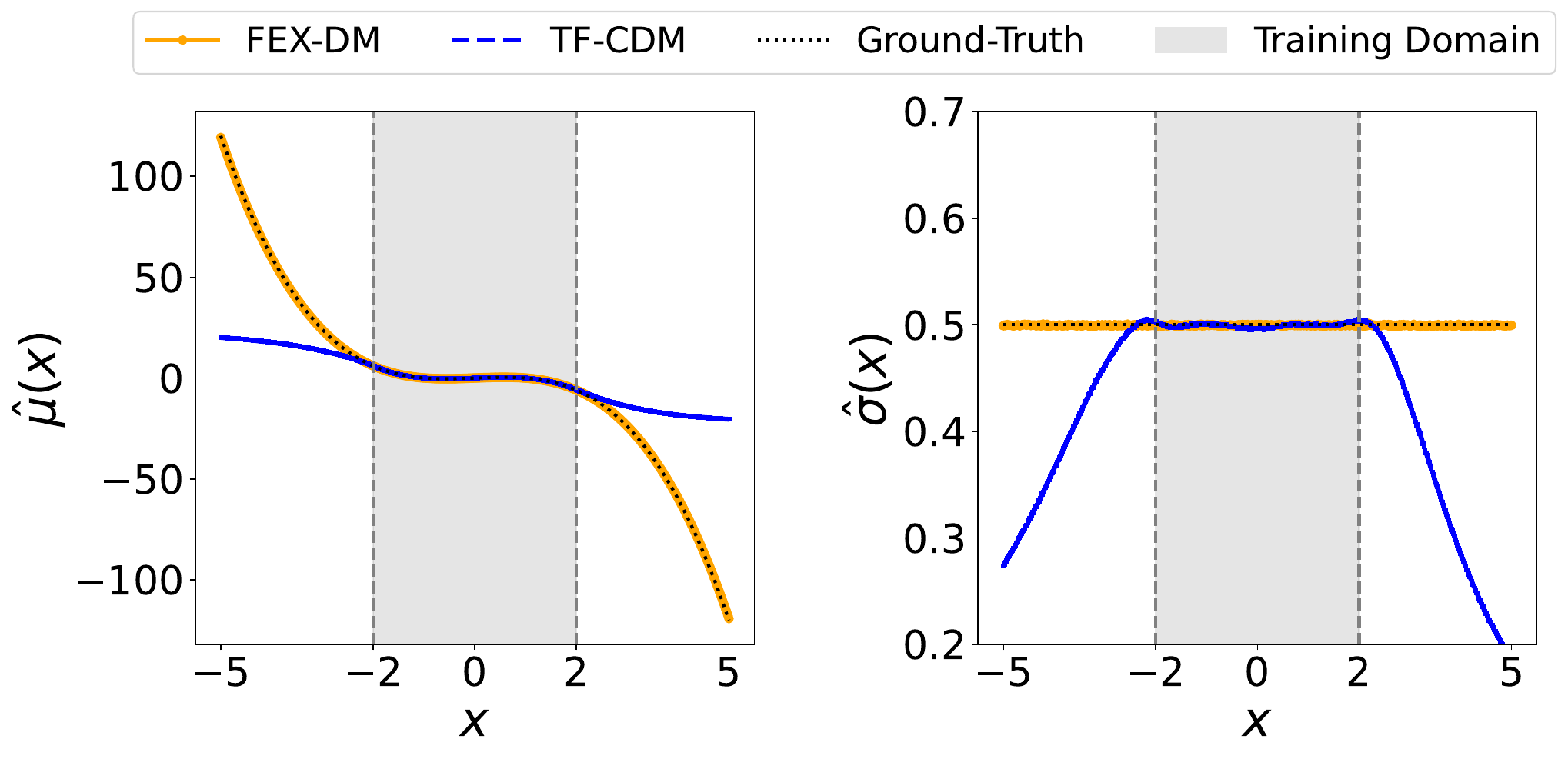}
    \caption{
    Comparison of the drift and diffusion functions learned by FEX-DM and TF-CDM against the ground truth over the domain $[-5,5]$ for the SDE with double well potential.  Left: drift term $\hat{\mu}(x)$; Right: diffusion term $\hat{\sigma}(x)$; The shaded region indicates the training domain $[-2,2]$.}\label{Fig3Dwell}
\end{figure}

After the second-stage training, we randomly select 500,000 trajectories with initial conditions $\mathbf{x}_0 = -5,1.5,5$, covering both in-domain and out-of-domain values to evaluate the framework's ability to predict SDE dynamics over \(T=1\).
Figure \ref{Fig1Dwell} compares the predicted mean trajectories and uncertainty bands, while Figure \ref{Fig2DwellExtended} shows the corresponding one-step conditional distributions. In both views, FEX-DM consistently aligns better with the ground truth than TF-CDM, accurately capturing mean, variance, and full conditional structure, even for out-of-domain inputs where TF-CDM deviates significantly in uncertainty and distributional shape. 
 Figure~\ref{fig:DwellEvul} further shows the corresponding conditional density evolution for $\mathbf{x}_0 = 1.5$ at selected times $t = 5,\ 10,\ 30,\ 100$, capturing the emergence of bimodality over time. Then we test the effective drift \(\hat{\mu}\) and \(\hat{\sigma}\) estimated by our framework in the extended
domain \([-5,5]\), as shown in Figure \ref{Fig3Dwell}. This further demonstrates the strong generalizability of FEX-DM.

\subsubsection{2-dimensional OL Process} We test our framework on a two-dimensional OL process governed by the following:
\begin{equation}\label{OLSDE}
   d\mathbf{x}_t = -\nabla V(\mathbf{x}_t) dt + \Sigma dW_t,
\end{equation}
where the state variable is
\(
\mathbf{x}_t = ({x}_{1} , {x}_{2} )\in \mathbb{R}^2.
\)
And the drift term is derived from the potential function:
\[
V(\mathbf{x}_t) = 2.5({x}_{1}^2 - 1)^2 + 5{x}_{2}^2,
\]
with diffusion function by:
\[
\Sigma = \begin{pmatrix}
    \sqrt{2} & 0 \\
    0 & \sqrt{2}
\end{pmatrix}.
\]
We simulate $35,000$ trajectories with initial values uniformly sampled from $(-1.5,1.5)\times(-1,1)$ and integrated up to $T=1.0$. After 200 iterations for training in the first stage, the controller NN identifies an operator sequence identical to that of the true drift function. Although the identified expression includes additional higher-order terms, their coefficients are relatively small and can be treated as residual errors, which are handled in the second stage of the FEX-DM framework. Combined with candidate optimization, this generates the best score expression for $\hat{D}(\mathbf{x}_t)$, as shown in Table \ref{OL2Tab}.  It indicates that the deterministic component has been successfully captured by FEX.

\begin{table}[!ht]
    \centering
    \renewcommand{\arraystretch}{1.5}
    \setlength{\tabcolsep}{10pt} % Adjust column spacing
    \small % Reduce font size if needed
    \caption{Comparison of true drift term and the FEX-learned deterministic component.}\label{OL2Tab}
    \begin{tabular}{|c|c|}
        \hline
        \textbf{True Drift Expression (2D)} & \textbf{Best Approximation for $\hat{D}(\mathbf{x}_t)$ (2D)} \\
        \hline
        $-10{x}_{1} ({x}_{1}^2-1)$ & $-9.9178{x}_{1}^3 + 0.1625{x}_{2}^3$ \\
         & $+9.8165{x}_{1} + 0.1204 {x}_{2} +0.03$ \\

        \hline
        $-10 {x}_{2}$ &
         $-0.0613 {x}_{1} - 9.9911 {x}_{2}+0.0011$ \\
        \hline
    \end{tabular}

\end{table}

\begin{figure}[!htbp]
    \centering
    \includegraphics[width=1\linewidth]{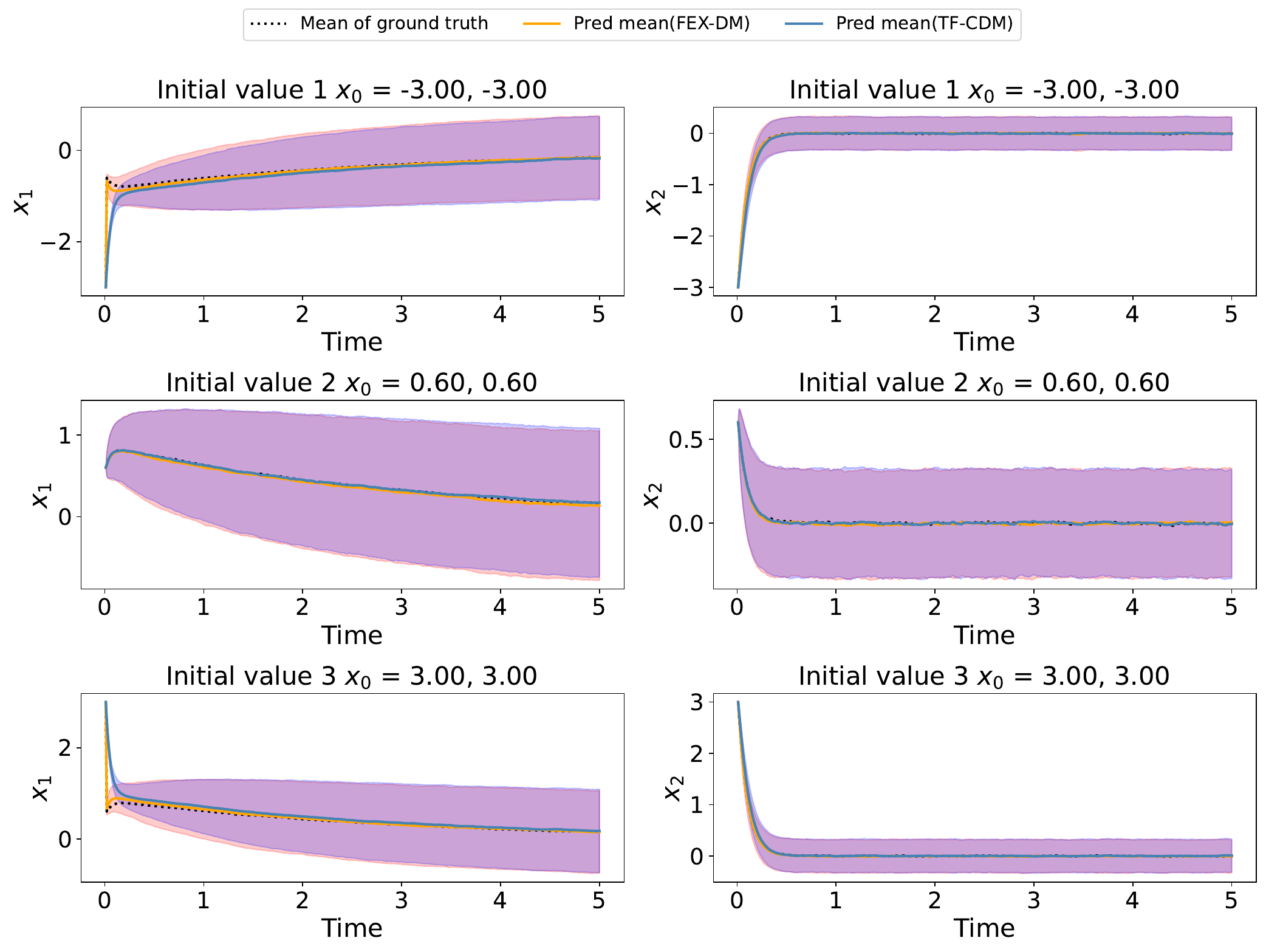}
    \caption{Comparison of the mean and the standard deviation of solutions for $\mathbf{x}_0=(-3,-3), (0.6,0.6)$, and $(3,3)$, obtained by the FEX-DM, TF-CDM and the ground truth for SDE with OL process.} \label{Fig1OL}
\end{figure}

\begin{figure}[!htbp]
    \centering
    \includegraphics[width=0.65\linewidth]{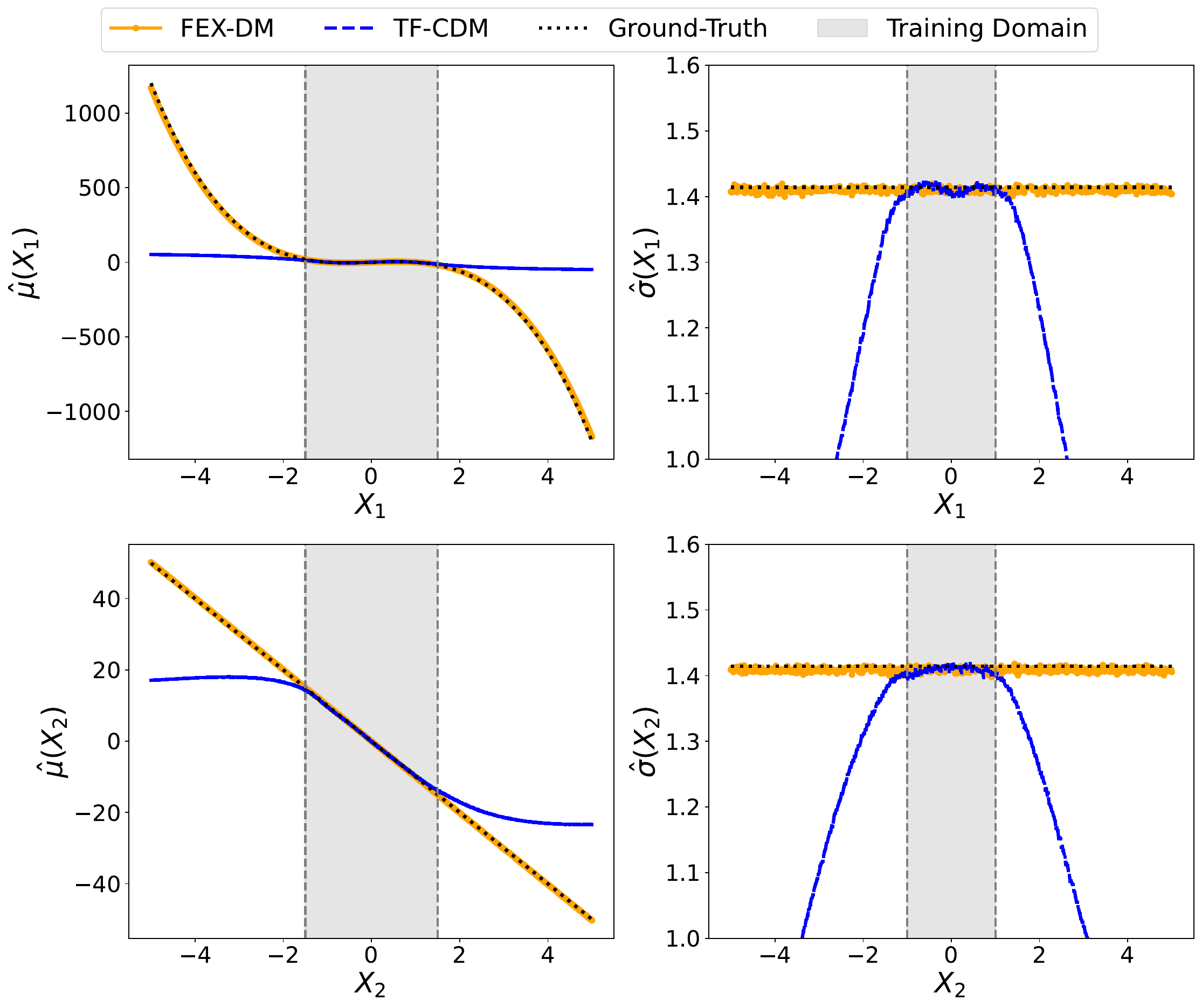}
    \caption{Comparison of drift and diffusion functions obtained by the FEX-DM and the ground truth over domain $[-5,5]$ for the SDE with OL process. Here $\hat{u}(\mathbf{x}_1)$ and $\hat{u}(\mathbf{x}_2)$ are the expressions in the second column of Table \ref{OL2Tab}. $\hat{\sigma}(\mathbf{x}_1)$ and $\hat{\sigma}(\mathbf{x}_2)$ are $\sqrt{2}$ in this case.}\label{Fig3OL}
\end{figure}

After the second-stage training, we randomly select 65,000 trajectories with initial conditions $\mathbf{x}_0 = (-3,-3), (0.6,0.6)$ and $(3,3)$, covering both in-domain and out-of-domain values to evaluate the framework's ability to predict SDE dynamics over \(T=5\).
Figure \eqref{Fig1OL} illustrates how the predicted means and standard deviations of the solutions for these initial conditions align with the ground truth. Although training was limited to $T=1.0$, the model generalizes well, maintaining close agreement with the exact solutions throughout the extended time horizon. Furthermore, 
 we also compare their performance in an extended domain $[-5,5] \times [-5,5]$ in Figure \ref{Fig3OL}. The observation is that the learned functions remain accurate within the training interval and continue to behave consistently in extrapolated regions.

\subsection{SDEs with Non-Gaussian Noise}\label{secNonGaussian}
SDEs driven by non-Gaussian noise model systems with irregular, discontinuous, or burst-like fluctuations and are widely applied in finance \cite{cont2003financial}, ecology \cite{katul2005mechanistic}, and neuroscience \cite{lindner2001transmission}. Here we test our framework on a one-dimensional SDE driven by exponential noise, formulated as:
\begin{equation}\label{expSDE}
    d\mathbf{x}_t = \mu \mathbf{x}_tdt+\sigma\sqrt{dt}\eta_t, \quad \eta_t \sim \text{exp}(1),
\end{equation}
where $\eta_t$ has an exponential PDF $f_{\eta}(x) = e^{-x}, x \ge 0$, and $\mu=-2.0, \sigma = 0.1$. We generate $L=10,000$ trajectories samples with initial values uniformly sampled from $(0,2.5)$ to train the two-stage framework. {When applying the formulation in \eqref{CO} to non-Gaussian noise, such as exponential noise, it is necessary to subtract the mean of the noise distribution to ensure zero-mean forcing.} After 200 training iterations in the first stage, the controller NN identifies an operator sequence identical to that of the true drift function. Combined with candidate optimization, this generates the best score expression for $\hat{D}(\mathbf{x}_t)$, as shown in Table \ref{ExpTab}.  It indicates that the deterministic component has been successfully captured by FEX.
\begin{table}[!ht]
\centering
\caption{Comparison of true drift term and the FEX-learned deterministic component.} \label{ExpTab}
\begin{tabular}{| >{\centering\arraybackslash}m{6cm} |>{\centering\arraybackslash}m{6cm} |}
\hline
\textbf{True drift expression} &
\textbf{Best expression for $\hat{D}(\mathbf{x}_t)$} \\ \hline
 $- 2\mathbf{x}_t$ &
$-1.9750528\mathbf{x}_t$
 \\ \hline
\end{tabular}

\end{table}

\begin{figure}[!htbp]
    \centering
    \includegraphics[width=0.9\linewidth]{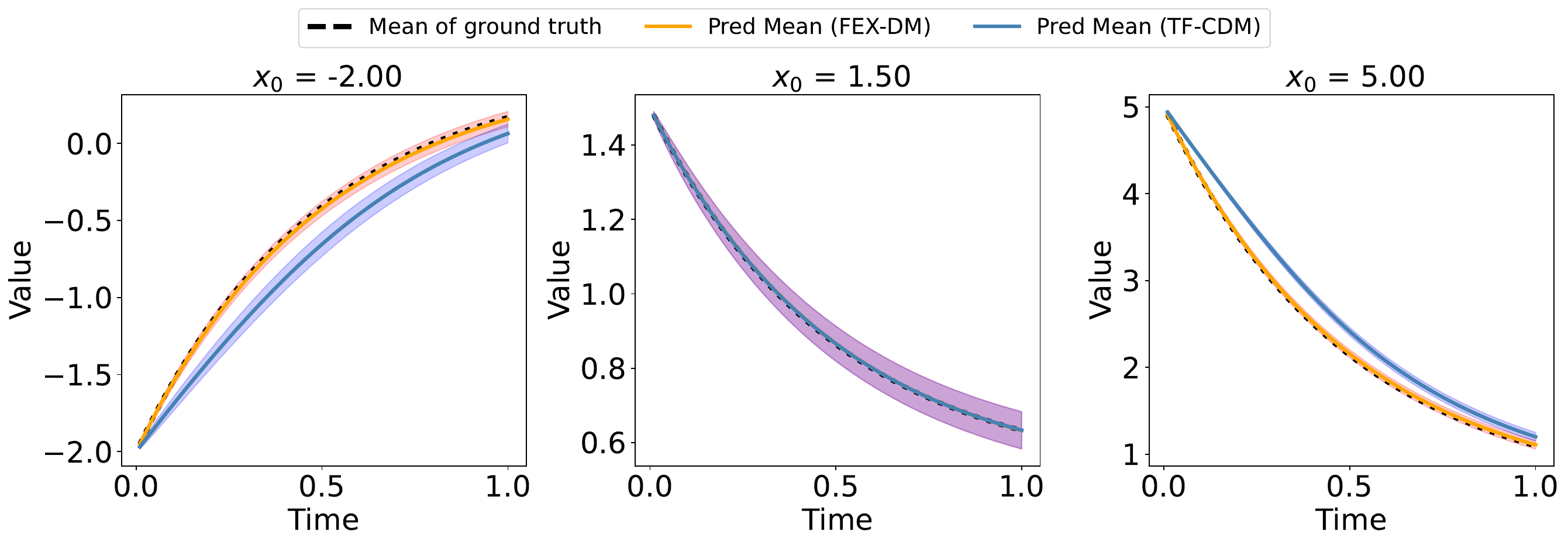}
    \caption{
Comparison of predicted mean trajectories and corresponding mean $\pm$ standard deviation bands for solutions to the SDE with exponential diffusion, evaluated at initial conditions $\mathbf{x}_0=-2, 1.5, 5$, obtained using the FEX-DM, TF-CDM, and the ground truth.} \label{Fig1Exp}
\end{figure}

\begin{figure}[!htbp]
    \centering
    \includegraphics[width=0.9\linewidth]{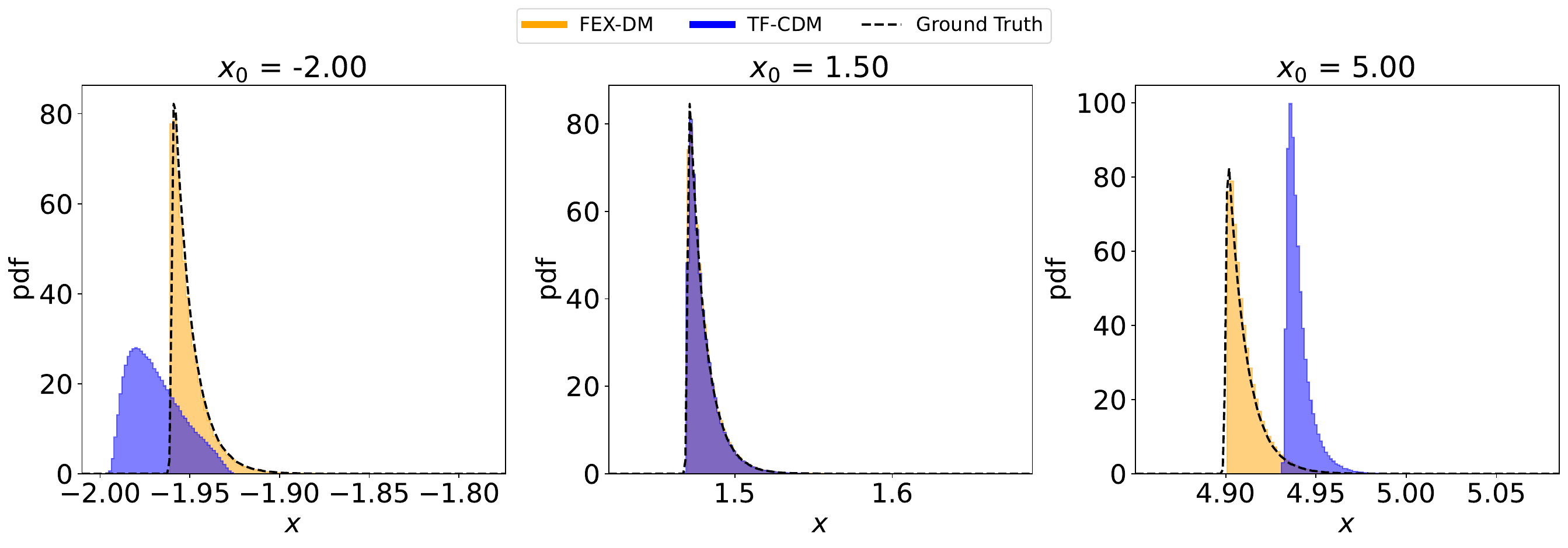}
    \caption{Comparison of conditional PDF $p_{\mathbf{x}_{t+\Delta t|\mathbf{x}_t}}(\mathbf{x}_{t+\Delta t}|\mathbf{x}_t = c)$ estimated by FEX-DM, TF-CDM, and the ground truth for the SDE with exponential diffusion, with $c=-2, 1.5, 5$ at $t=0$, respectively.}\label{Fig2Exp}
\end{figure}

\begin{figure}[!htbp]
    \centering
    \includegraphics[width=0.65\linewidth]{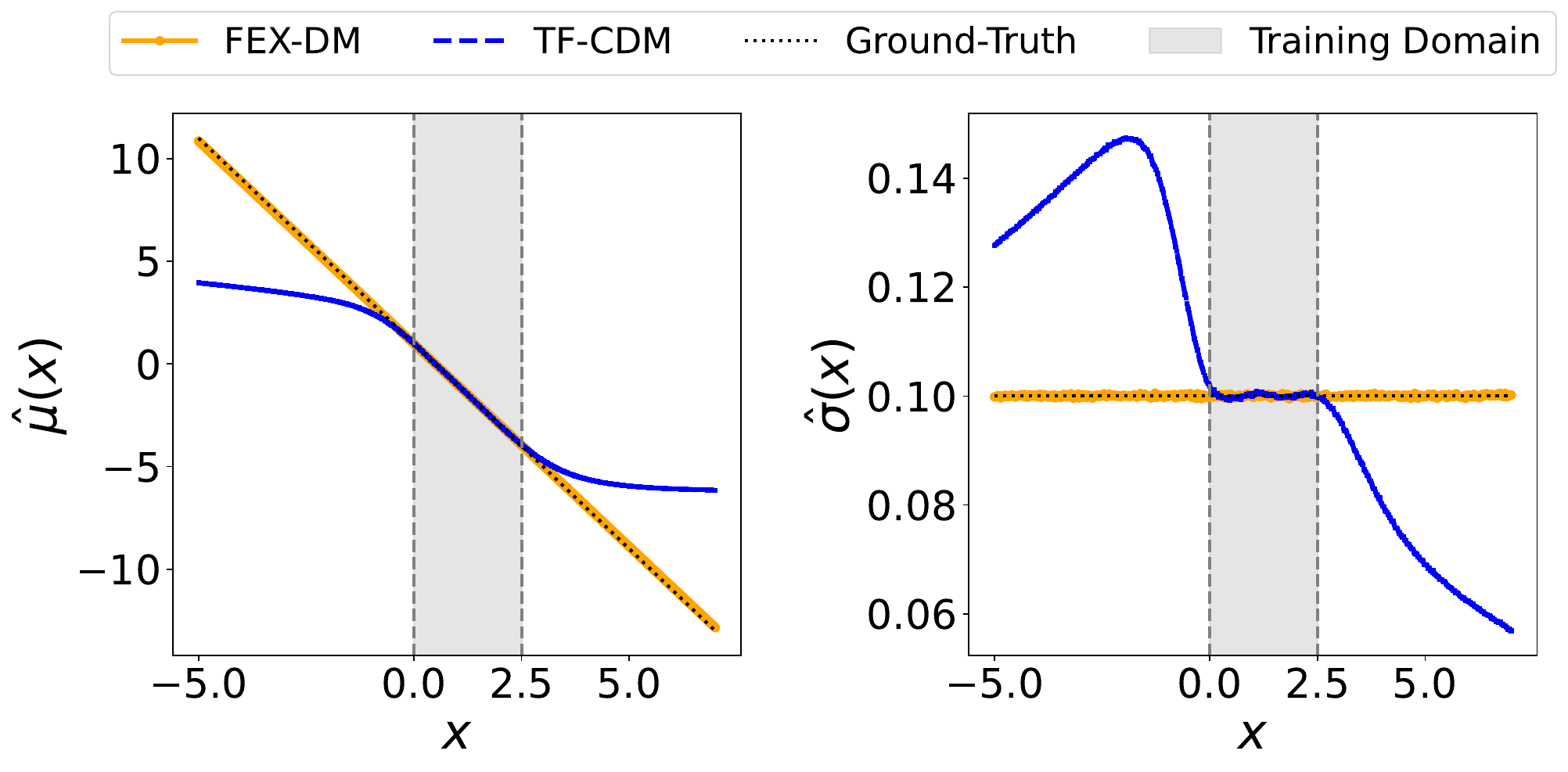}
    \caption{Comparison of the drift and diffusion functions learned by FEX-DM, TF-CDM and the ground truth over the domain $[-5,7]$ for the SDE with exponential diffusion. Left: drift term $\hat{\mu}(x)$; Right: diffusion term $\hat{\sigma}(x)$; The shaded region indicates the training domain $(0,2.5)$.}\label{Fig3Exp}
\end{figure}

After the second-stage training, we randomly select 500,000 trajectories with initial conditions $\mathbf{x}_0 = -2,1.5,5$, covering both in-domain and out-of-domain values to evaluate the framework's ability to predict SDE dynamics over \(T=1\).
Figure \ref{Fig1Exp} compares the predicted mean trajectories and uncertainty bands, while Figure \ref{Fig2Exp} shows the corresponding one-step conditional distributions. In both views, FEX-DM consistently aligns better with the ground truth than TF-CDM, accurately capturing mean, variance, and full conditional structure, even for out-of-domain inputs where TF-CDM deviates significantly in uncertainty and distributional shape. In addition, we test the effective drift \(\hat{\mu}\) and \(\hat{\sigma}\) estimated by our framework in the extended
domain \([-5,7]\), as shown in Figure \ref{Fig3Exp}. This further demonstrates the strong generalizability of FEX-DM.

\section{Discussion and Perspectives}\label{sec:con} 
This paper introduces a novel two-stage framework for modeling stochastic systems. The first stage learns the deterministic component using FEX to derive explicit mathematical expressions. The second stage complements this with a generative method for modeling the stochastic component. The framework prioritizes transparency and interpretability, avoiding dependence on black-box neural architectures. By learning closed-form expressions directly from the data, the model demonstrates strong predictive performance across a wide range of stochastic systems. It generalizes well beyond the training domain, accurately captures both short- and long-term dynamics, and effectively models complex behaviors in linear, nonlinear, and multidimensional settings.

Despite these promising results, several extensions could further improve and broaden the approach. First, incorporating state-dependent or control-dependent diffusion functions would allow the model to represent more complex stochastic component, including those exhibiting nonlocal behavior or critical transitions. Second, extending the framework to accommodate non-autonomous systems, where system properties evolve explicitly over time, would expand its applicability to time-varying environments. Third, adapting the model to handle more advanced stochastic structures, such as jump processes or systems driven by fractional Brownian motion, could increase its utility in domains like finance, quantum physics~\cite{zhang2025mitigation}, and climate science~\cite{mou2023combining}. Finally, a deeper theoretical investigation of the underlying mechanisms of the score-based sampling strategy, particularly its convergence properties and error bounds, would provide stronger guarantees and insight into the foundations of the method. These directions highlight the potential for continued development of mathematically grounded and interpretable approaches for learning complex stochastic systems.

\section{Acknowledgements}
The research of Senwei Liang was partially supported by the U.S. Department of Energy, Office of Science, Office of Advanced Scientific Computing Research SciDAC program under Contract No. DE-AC02-05CH11231.  The research of Chunmei Wang was partially supported by National Science Foundation Grant DMS-2206332.  
\bibliographystyle{plain}
\bibliography{main}

\begin{thebibliography}{10}

\bibitem{archambeau2007gaussian}
Cedric Archambeau, Dan Cornford, Manfred Opper, and John Shawe-Taylor.
\newblock Gaussian process approximations of stochastic differential equations.
\newblock In {\em Gaussian Processes in Practice}, pages 1--16. PMLR, 2007.

\bibitem{bose2009stochastic}
Thomas Bose and Steffen Trimper.
\newblock Stochastic model for tumor growth with immunization.
\newblock {\em Physical Review E—Statistical, Nonlinear, and Soft Matter
  Physics}, 79(5):051903, 2009.

\bibitem{brunton2016discovering}
Steven~L Brunton, Joshua~L Proctor, and J~Nathan Kutz.
\newblock Discovering governing equations from data by sparse identification of
  nonlinear dynamical systems.
\newblock {\em Proceedings of the national academy of sciences},
  113(15):3932--3937, 2016.

\bibitem{cao2023genetic}
Lulu Cao, Zimo Zheng, Chenwen Ding, Jinkai Cai, and Min Jiang.
\newblock Genetic programming symbolic regression with simplification-pruning
  operator for solving differential equations.
\newblock In {\em International Conference on Neural Information Processing},
  pages 287--298. Springer, 2023.

\bibitem{chen2023data}
Xiaoli Chen, Jinqiao Duan, Jianyu Hu, and Dongfang Li.
\newblock Data-driven method to learn the most probable transition pathway and
  stochastic differential equation.
\newblock {\em Physica D: Nonlinear Phenomena}, 443:133559, 2023.

\bibitem{chen2024learning}
Yuan Chen and Dongbin Xiu.
\newblock Learning stochastic dynamical system via flow map operator.
\newblock {\em Journal of Computational Physics}, 508:112984, 2024.

\bibitem{chen2024modeling}
Yuan Chen and Dongbin Xiu.
\newblock Modeling unknown stochastic dynamical system subject to external
  excitation.
\newblock {\em arXiv preprint arXiv:2406.15747}, 2024.

\bibitem{chong2020vortices}
Kai~Leong Chong, Jun-Qiang Shi, Guang-Yu Ding, Shan-Shan Ding, Hao-Yuan Lu,
  Jin-Qiang Zhong, and Ke-Qing Xia.
\newblock Vortices as brownian particles in turbulent flows.
\newblock {\em Science advances}, 6(34):eaaz1110, 2020.

\bibitem{coffey2012langevin}
William Coffey and Yu~P Kalmykov.
\newblock {\em The Langevin equation: with applications to stochastic problems
  in physics, chemistry and electrical engineering}, volume~27.
\newblock World Scientific, 2012.

\bibitem{cont2003financial}
Rama Cont and Peter Tankov.
\newblock {\em Financial modelling with jump processes}.
\newblock Chapman and Hall/CRC, 2003.

\bibitem{cranmer2020discovering}
Miles Cranmer, Alvaro Sanchez~Gonzalez, Peter Battaglia, Rui Xu, Kyle Cranmer,
  David Spergel, and Shirley Ho.
\newblock Discovering symbolic models from deep learning with inductive biases.
\newblock {\em Advances in neural information processing systems},
  33:17429--17442, 2020.

\bibitem{darcy2023one}
Matthieu Darcy, Boumediene Hamzi, Giulia Livieri, Houman Owhadi, and Peyman
  Tavallali.
\newblock One-shot learning of stochastic differential equations with data
  adapted kernels.
\newblock {\em Physica D: Nonlinear Phenomena}, 444:133583, 2023.

\bibitem{du2024learning}
Jianda Du, Senwei Liang, and Chunmei Wang.
\newblock Learning epidemiological dynamics via the finite expression method.
\newblock {\em arXiv preprint arXiv:2412.21049}, 2024.

\bibitem{du2019gradient}
Simon Du, Jason Lee, Haochuan Li, Liwei Wang, and Xiyu Zhai.
\newblock Gradient descent finds global minima of deep neural networks.
\newblock In {\em International conference on machine learning}, pages
  1675--1685. PMLR, 2019.

\bibitem{einstein1905molekularkinetischen}
Albert Einstein.
\newblock {\"U}ber die von der molekularkinetischen theorie der w{\"a}rme
  geforderte bewegung von in ruhenden fl{\"u}ssigkeiten suspendierten teilchen.
\newblock {\em Annalen der physik}, 4, 1905.

\bibitem{elgin1984fokker}
JN~Elgin.
\newblock The fokker-planck equation: methods of solution and applications.
\newblock {\em Optica Acta: International Journal of Optics},
  31(11):1206--1207, 1984.

\bibitem{fasel2021sindy}
Urban Fasel, Eurika Kaiser, J~Nathan Kutz, Bingni~W Brunton, and Steven~L
  Brunton.
\newblock Sindy with control: A tutorial.
\newblock In {\em 2021 60th IEEE conference on decision and control (CDC)},
  pages 16--21. IEEE, 2021.

\bibitem{fletcher2000practical}
Roger Fletcher.
\newblock {\em Practical methods of optimization}.
\newblock John Wiley \& Sons, 2000.

\bibitem{forrest1993genetic}
Stephanie Forrest.
\newblock Genetic algorithms: principles of natural selection applied to
  computation.
\newblock {\em Science}, 261(5123):872--878, 1993.

\bibitem{gardinerstochastic}
C~Gardiner.
\newblock Stochastic methods: A handbook for the natural and social sciences
  fourth edition (2009).

\bibitem{gardiner2009stochastic}
CW~Gardiner.
\newblock Handbook of stochastic methods, eds. 4th, 2009.

\bibitem{gower2019sgd}
Robert~Mansel Gower, Nicolas Loizou, Xun Qian, Alibek Sailanbayev, Egor
  Shulgin, and Peter Richt{\'a}rik.
\newblock Sgd: General analysis and improved rates.
\newblock In {\em International conference on machine learning}, pages
  5200--5209. PMLR, 2019.

\bibitem{hanggi1990reaction}
Peter H{\"a}nggi, Peter Talkner, and Michal Borkovec.
\newblock Reaction-rate theory: fifty years after kramers.
\newblock {\em Reviews of modern physics}, 62(2):251, 1990.

\bibitem{he2019bag}
Tong He, Zhi Zhang, Hang Zhang, Zhongyue Zhang, Junyuan Xie, and Mu~Li.
\newblock Bag of tricks for image classification with convolutional neural
  networks.
\newblock In {\em Proceedings of the IEEE/CVF conference on computer vision and
  pattern recognition}, pages 558--567, 2019.

\bibitem{jiang2023symbolic}
Nan Jiang and Yexiang Xue.
\newblock Symbolic regression via control variable genetic programming.
\newblock In {\em Joint European Conference on Machine Learning and Knowledge
  Discovery in Databases}, pages 178--195. Springer, 2023.

\bibitem{jiang2023finite}
Zhongyi Jiang, Chunmei Wang, and Haizhao Yang.
\newblock Finite expression methods for discovering physical laws from data.
\newblock {\em arXiv preprint arXiv:2305.08342}, 2023.

\bibitem{kamienny2022end}
Pierre-Alexandre Kamienny, St{\'e}phane d'Ascoli, Guillaume Lample, and
  Fran{\c{c}}ois Charton.
\newblock End-to-end symbolic regression with transformers.
\newblock {\em Advances in Neural Information Processing Systems},
  35:10269--10281, 2022.

\bibitem{karatzas1991brownian}
Ioannis Karatzas and Steven Shreve.
\newblock {\em Brownian motion and stochastic calculus}, volume 113.
\newblock Springer Science \& Business Media, 1991.

\bibitem{kariya2003options}
Takeaki Kariya, Regina~Y Liu, Takeaki Kariya, and Regina~Y Liu.
\newblock Options, futures and other derivatives.
\newblock {\em Asset Pricing: -Discrete Time Approach-}, pages 9--26, 2003.

\bibitem{katul2005mechanistic}
GABRIEL~GEORGE Katul, Amilcare Porporato, R~Nathan, M~Siqueira, MB~Soons,
  Davide Poggi, HS~Horn, and SA~Levin.
\newblock Mechanistic analytical models for long-distance seed dispersal by
  wind.
\newblock {\em The American Naturalist}, 166(3):368--381, 2005.

\bibitem{kramers1940brownian}
Hendrik~Anthony Kramers.
\newblock Brownian motion in a field of force and the diffusion model of
  chemical reactions.
\newblock {\em physica}, 7(4):284--304, 1940.

\bibitem{li2022transformer}
Wenqiang Li, Weijun Li, Linjun Sun, Min Wu, Lina Yu, Jingyi Liu, Yanjie Li, and
  Songsong Tian.
\newblock Transformer-based model for symbolic regression via joint supervised
  learning.
\newblock In {\em The Eleventh International Conference on Learning
  Representations}, 2022.

\bibitem{li2021data}
Yang Li and Jinqiao Duan.
\newblock A data-driven approach for discovering stochastic dynamical systems
  with non-gaussian l{\'e}vy noise.
\newblock {\em Physica D: Nonlinear Phenomena}, 417:132830, 2021.

\bibitem{liang2022finite}
Senwei Liang and Haizhao Yang.
\newblock Finite expression method for solving high-dimensional partial
  differential equations.
\newblock {\em arXiv preprint arXiv:2206.10121}, 2022.

\bibitem{lindner2001transmission}
Benjamin Lindner and Lutz Schimansky-Geier.
\newblock Transmission of noise coded versus additive signals through a
  neuronal ensemble.
\newblock {\em Physical review letters}, 86(14):2934, 2001.

\bibitem{liu2024training}
Yanfang Liu, Yuan Chen, Dongbin Xiu, and Guannan Zhang.
\newblock A training-free conditional diffusion model for learning stochastic
  dynamical systems.
\newblock {\em arXiv preprint arXiv:2410.03108}, 2024.

\bibitem{liu2024kan}
Ziming Liu, Yixuan Wang, Sachin Vaidya, Fabian Ruehle, James Halverson, Marin
  Solja{\v{c}}i{\'c}, Thomas~Y Hou, and Max Tegmark.
\newblock Kan: Kolmogorov-arnold networks.
\newblock {\em arXiv preprint arXiv:2404.19756}, 2024.

\bibitem{montanelli2020error}
Hadrien Montanelli and Haizhao Yang.
\newblock Error bounds for deep relu networks using the kolmogorov--arnold
  superposition theorem.
\newblock {\em Neural Networks}, 129:1--6, 2020.

\bibitem{mou2023efficient}
Changhong Mou, Nan Chen, and Traian Iliescu.
\newblock An efficient data-driven multiscale stochastic reduced order modeling
  framework for complex systems.
\newblock {\em Journal of Computational Physics}, 493:112450, 2023.

\bibitem{mou2023combining}
Changhong Mou, Leslie~M Smith, and Nan Chen.
\newblock Combining stochastic parameterized reduced-order models with machine
  learning for data assimilation and uncertainty quantification with partial
  observations.
\newblock {\em Journal of Advances in Modeling Earth Systems},
  15(10):e2022MS003597, 2023.

\bibitem{oh2023genetic}
Hongsup Oh, Roman Amici, Geoffrey Bomarito, Shandian Zhe, Robert Kirby, and
  Jacob Hochhalter.
\newblock Genetic programming based symbolic regression for analytical
  solutions to differential equations.
\newblock {\em arXiv preprint arXiv:2302.03175}, 2023.

\bibitem{oksendal2013stochastic}
Bernt Oksendal.
\newblock {\em Stochastic differential equations: an introduction with
  applications}.
\newblock Springer Science \& Business Media, 2013.

\bibitem{opper2019variational}
Manfred Opper.
\newblock Variational inference for stochastic differential equations.
\newblock {\em Annalen der Physik}, 531(3):1800233, 2019.

\bibitem{petersen2019deep}
Brenden~K Petersen, Mikel Landajuela, T~Nathan Mundhenk, Claudio~P Santiago,
  Soo~K Kim, and Joanne~T Kim.
\newblock Deep symbolic regression: Recovering mathematical expressions from
  data via risk-seeking policy gradients.
\newblock {\em arXiv preprint arXiv:1912.04871}, 2019.

\bibitem{qin2019data}
Tong Qin, Kailiang Wu, and Dongbin Xiu.
\newblock Data driven governing equations approximation using deep neural
  networks.
\newblock {\em Journal of Computational Physics}, 395:620--635, 2019.

\bibitem{rigas2015diffusive}
Georgios Rigas, Aimee~S Morgans, RD~Brackston, and Jonathan~F Morrison.
\newblock Diffusive dynamics and stochastic models of turbulent axisymmetric
  wakes.
\newblock {\em Journal of Fluid Mechanics}, 778:R2, 2015.

\bibitem{salimans2016improved}
Tim Salimans, Ian Goodfellow, Wojciech Zaremba, Vicki Cheung, Alec Radford, and
  Xi~Chen.
\newblock Improved techniques for training gans.
\newblock {\em Advances in neural information processing systems}, 29, 2016.

\bibitem{song2024finite}
Zezheng Song, Chunmei Wang, and Haizhao Yang.
\newblock Finite expression method for learning dynamics on complex networks.
\newblock {\em arXiv preprint arXiv:2401.03092}, 2024.

\bibitem{sutton2018reinforcement}
Richard~S Sutton.
\newblock Reinforcement learning: An introduction.
\newblock {\em A Bradford Book}, 2018.

\bibitem{udrescu2020ai}
Silviu-Marian Udrescu and Max Tegmark.
\newblock Ai feynman: A physics-inspired method for symbolic regression.
\newblock {\em Science advances}, 6(16):eaay2631, 2020.

\bibitem{uhlenbeck1930theory}
George~E Uhlenbeck and Leonard~S Ornstein.
\newblock On the theory of the brownian motion.
\newblock {\em Physical review}, 36(5):823, 1930.

\bibitem{vasicek1977equilibrium}
Oldrich Vasicek.
\newblock An equilibrium characterization of the term structure.
\newblock {\em Journal of financial economics}, 5(2):177--188, 1977.

\bibitem{wei2024closed}
Shu Wei, Yanjie Li, Lina Yu, Min Wu, Weijun Li, Meilan Hao, Wenqiang Li, Jingyi
  Liu, and Yusong Deng.
\newblock Closed-form symbolic solutions: A new perspective on solving partial
  differential equations.
\newblock {\em arXiv preprint arXiv:2405.14620}, 2024.

\bibitem{werner2024sample}
Steffen~WR Werner and Benjamin Peherstorfer.
\newblock On the sample complexity of stabilizing linear dynamical systems from
  data.
\newblock {\em Foundations of Computational Mathematics}, 24(3):955--987, 2024.

\bibitem{xu2024modeling}
Zhongshu Xu, Yuan Chen, Qifan Chen, and Dongbin Xiu.
\newblock Modeling unknown stochastic dynamical system via autoencoder.
\newblock {\em Journal of Machine Learning for Modeling and Computing}, 5(3),
  2024.

\bibitem{yang2024pseudoreversible}
Minglei Yang, Pengjun Wang, Diego del Castillo-Negrete, Yanzhao Cao, and
  Guannan Zhang.
\newblock A pseudoreversible normalizing flow for stochastic dynamical systems
  with various initial distributions.
\newblock {\em SIAM Journal on Scientific Computing}, 46(4):C508--C533, 2024.

\bibitem{zhang2025mitigation}
Chengxi Zhang, Justin Phillips, Inder Monga, Erhan Saglamyurek, Qiming Wu, and
  Hartmut Haeffner.
\newblock Mitigation of birefringence in cavity-based quantum networks using
  frequency-encoded photons.
\newblock {\em Physical Review A}, 111(6):062608, 2025.

\bibitem{zhang2019convergence}
Linan Zhang and Hayden Schaeffer.
\newblock On the convergence of the sindy algorithm.
\newblock {\em Multiscale Modeling \& Simulation}, 17(3):948--972, 2019.

\bibitem{zhu1997algorithm}
Ciyou Zhu, Richard~H Byrd, Peihuang Lu, and Jorge Nocedal.
\newblock Algorithm 778: L-bfgs-b: Fortran subroutines for large-scale
  bound-constrained optimization.
\newblock {\em ACM Transactions on mathematical software (TOMS)},
  23(4):550--560, 1997.

\end{thebibliography}
\end{document}